\documentclass[conference, 10pt]{IEEEtran}
\ifCLASSINFOpdf
\else
\fi
\usepackage[cmex10]{amsmath}
\newcommand\scalemath[2]{\scalebox{#1}{\mbox{\ensuremath{\displaystyle #2}}}}
\usepackage{amsfonts}
\usepackage{subfigure}
\usepackage{color}
\definecolor{AliceBlue}{rgb}{0.94,0.97,1.00}
\definecolor{AntiqueWhite1}{rgb}{1.00,0.94,0.86}
\definecolor{AntiqueWhite2}{rgb}{0.93,0.87,0.80}
\definecolor{AntiqueWhite3}{rgb}{0.80,0.75,0.69}
\definecolor{AntiqueWhite4}{rgb}{0.55,0.51,0.47}
\definecolor{AntiqueWhite}{rgb}{0.98,0.92,0.84}
\definecolor{BlanchedAlmond}{rgb}{1.00,0.92,0.80}
\definecolor{BlueViolet}{rgb}{0.54,0.17,0.89}
\definecolor{CadetBlue1}{rgb}{0.60,0.96,1.00}
\definecolor{CadetBlue2}{rgb}{0.56,0.90,0.93}
\definecolor{CadetBlue3}{rgb}{0.48,0.77,0.80}
\definecolor{CadetBlue4}{rgb}{0.33,0.53,0.55}
\definecolor{CadetBlue}{rgb}{0.37,0.62,0.63}
\definecolor{CornflowerBlue}{rgb}{0.39,0.58,0.93}
\definecolor{DarkBlue}{rgb}{0.00,0.00,0.55}
\definecolor{DarkCyan}{rgb}{0.00,0.55,0.55}
\definecolor{DarkGoldenrod1}{rgb}{1.00,0.73,0.06}
\definecolor{DarkGoldenrod2}{rgb}{0.93,0.68,0.05}
\definecolor{DarkGoldenrod3}{rgb}{0.80,0.58,0.05}
\definecolor{DarkGoldenrod4}{rgb}{0.55,0.40,0.03}
\definecolor{DarkGoldenrod}{rgb}{0.72,0.53,0.04}
\definecolor{DarkGray}{rgb}{0.66,0.66,0.66}
\definecolor{DarkGreen}{rgb}{0.00,0.39,0.00}
\definecolor{DarkGrey}{rgb}{0.66,0.66,0.66}
\definecolor{DarkKhaki}{rgb}{0.74,0.72,0.42}
\definecolor{DarkMagenta}{rgb}{0.55,0.00,0.55}
\definecolor{DarkOliveGreen1}{rgb}{0.79,1.00,0.44}
\definecolor{DarkOliveGreen2}{rgb}{0.74,0.93,0.41}
\definecolor{DarkOliveGreen3}{rgb}{0.64,0.80,0.35}
\definecolor{DarkOliveGreen4}{rgb}{0.43,0.55,0.24}
\definecolor{DarkOliveGreen}{rgb}{0.33,0.42,0.18}
\definecolor{DarkOrange1}{rgb}{1.00,0.50,0.00}
\definecolor{DarkOrange2}{rgb}{0.93,0.46,0.00}
\definecolor{DarkOrange3}{rgb}{0.80,0.40,0.00}
\definecolor{DarkOrange4}{rgb}{0.55,0.27,0.00}
\definecolor{DarkOrange}{rgb}{1.00,0.55,0.00}
\definecolor{DarkOrchid1}{rgb}{0.75,0.24,1.00}
\definecolor{DarkOrchid2}{rgb}{0.70,0.23,0.93}
\definecolor{DarkOrchid3}{rgb}{0.60,0.20,0.80}
\definecolor{DarkOrchid4}{rgb}{0.41,0.13,0.55}
\definecolor{DarkOrchid}{rgb}{0.60,0.20,0.80}
\definecolor{DarkRed}{rgb}{0.55,0.00,0.00}
\definecolor{DarkSalmon}{rgb}{0.91,0.59,0.48}
\definecolor{DarkSeaGreen1}{rgb}{0.76,1.00,0.76}
\definecolor{DarkSeaGreen2}{rgb}{0.71,0.93,0.71}
\definecolor{DarkSeaGreen3}{rgb}{0.61,0.80,0.61}
\definecolor{DarkSeaGreen4}{rgb}{0.41,0.55,0.41}
\definecolor{DarkSeaGreen}{rgb}{0.56,0.74,0.56}
\definecolor{DarkSlateBlue}{rgb}{0.28,0.24,0.55}
\definecolor{DarkSlateGray1}{rgb}{0.59,1.00,1.00}
\definecolor{DarkSlateGray2}{rgb}{0.55,0.93,0.93}
\definecolor{DarkSlateGray3}{rgb}{0.47,0.80,0.80}
\definecolor{DarkSlateGray4}{rgb}{0.32,0.55,0.55}
\definecolor{DarkSlateGray}{rgb}{0.18,0.31,0.31}
\definecolor{DarkSlateGrey}{rgb}{0.18,0.31,0.31}
\definecolor{DarkTurquoise}{rgb}{0.00,0.81,0.82}
\definecolor{DarkViolet}{rgb}{0.58,0.00,0.83}
\definecolor{DeepPink1}{rgb}{1.00,0.08,0.58}
\definecolor{DeepPink2}{rgb}{0.93,0.07,0.54}
\definecolor{DeepPink3}{rgb}{0.80,0.06,0.46}
\definecolor{DeepPink4}{rgb}{0.55,0.04,0.31}
\definecolor{DeepPink}{rgb}{1.00,0.08,0.58}
\definecolor{DeepSkyBlue1}{rgb}{0.00,0.75,1.00}
\definecolor{DeepSkyBlue2}{rgb}{0.00,0.70,0.93}
\definecolor{DeepSkyBlue3}{rgb}{0.00,0.60,0.80}
\definecolor{DeepSkyBlue4}{rgb}{0.00,0.41,0.55}
\definecolor{DeepSkyBlue}{rgb}{0.00,0.75,1.00}
\definecolor{DimGray}{rgb}{0.41,0.41,0.41}
\definecolor{DimGrey}{rgb}{0.41,0.41,0.41}
\definecolor{DodgerBlue1}{rgb}{0.12,0.56,1.00}
\definecolor{DodgerBlue2}{rgb}{0.11,0.53,0.93}
\definecolor{DodgerBlue3}{rgb}{0.09,0.45,0.80}
\definecolor{DodgerBlue4}{rgb}{0.06,0.31,0.55}
\definecolor{DodgerBlue}{rgb}{0.12,0.56,1.00}
\definecolor{FloralWhite}{rgb}{1.00,0.98,0.94}
\definecolor{ForestGreen}{rgb}{0.13,0.55,0.13}
\definecolor{GhostWhite}{rgb}{0.97,0.97,1.00}
\definecolor{GreenYellow}{rgb}{0.68,1.00,0.18}
\definecolor{HotPink1}{rgb}{1.00,0.43,0.71}
\definecolor{HotPink2}{rgb}{0.93,0.42,0.65}
\definecolor{HotPink3}{rgb}{0.80,0.38,0.56}
\definecolor{HotPink4}{rgb}{0.55,0.23,0.38}
\definecolor{HotPink}{rgb}{1.00,0.41,0.71}
\definecolor{IndianRed1}{rgb}{1.00,0.42,0.42}
\definecolor{IndianRed2}{rgb}{0.93,0.39,0.39}
\definecolor{IndianRed3}{rgb}{0.80,0.33,0.33}
\definecolor{IndianRed4}{rgb}{0.55,0.23,0.23}
\definecolor{IndianRed}{rgb}{0.80,0.36,0.36}
\definecolor{LavenderBlush1}{rgb}{1.00,0.94,0.96}
\definecolor{LavenderBlush2}{rgb}{0.93,0.88,0.90}
\definecolor{LavenderBlush3}{rgb}{0.80,0.76,0.77}
\definecolor{LavenderBlush4}{rgb}{0.55,0.51,0.53}
\definecolor{LavenderBlush}{rgb}{1.00,0.94,0.96}
\definecolor{LawnGreen}{rgb}{0.49,0.99,0.00}
\definecolor{LemonChiffon1}{rgb}{1.00,0.98,0.80}
\definecolor{LemonChiffon2}{rgb}{0.93,0.91,0.75}
\definecolor{LemonChiffon3}{rgb}{0.80,0.79,0.65}
\definecolor{LemonChiffon4}{rgb}{0.55,0.54,0.44}
\definecolor{LemonChiffon}{rgb}{1.00,0.98,0.80}
\definecolor{LightBlue1}{rgb}{0.75,0.94,1.00}
\definecolor{LightBlue2}{rgb}{0.70,0.87,0.93}
\definecolor{LightBlue3}{rgb}{0.60,0.75,0.80}
\definecolor{LightBlue4}{rgb}{0.41,0.51,0.55}
\definecolor{LightBlue}{rgb}{0.68,0.85,0.90}
\definecolor{LightCoral}{rgb}{0.94,0.50,0.50}
\definecolor{LightCyan1}{rgb}{0.88,1.00,1.00}
\definecolor{LightCyan2}{rgb}{0.82,0.93,0.93}
\definecolor{LightCyan3}{rgb}{0.71,0.80,0.80}
\definecolor{LightCyan4}{rgb}{0.48,0.55,0.55}
\definecolor{LightCyan}{rgb}{0.88,1.00,1.00}
\definecolor{LightGoldenrod1}{rgb}{1.00,0.93,0.55}
\definecolor{LightGoldenrod2}{rgb}{0.93,0.86,0.51}
\definecolor{LightGoldenrod3}{rgb}{0.80,0.75,0.44}
\definecolor{LightGoldenrod4}{rgb}{0.55,0.51,0.30}
\definecolor{LightGoldenrodYellow}{rgb}{0.98,0.98,0.82}
\definecolor{LightGoldenrod}{rgb}{0.93,0.87,0.51}
\definecolor{LightGray}{rgb}{0.83,0.83,0.83}
\definecolor{LightGreen}{rgb}{0.56,0.93,0.56}
\definecolor{LightGrey}{rgb}{0.83,0.83,0.83}
\definecolor{LightPink1}{rgb}{1.00,0.68,0.73}
\definecolor{LightPink2}{rgb}{0.93,0.64,0.68}
\definecolor{LightPink3}{rgb}{0.80,0.55,0.58}
\definecolor{LightPink4}{rgb}{0.55,0.37,0.40}
\definecolor{LightPink}{rgb}{1.00,0.71,0.76}
\definecolor{LightSalmon1}{rgb}{1.00,0.63,0.48}
\definecolor{LightSalmon2}{rgb}{0.93,0.58,0.45}
\definecolor{LightSalmon3}{rgb}{0.80,0.51,0.38}
\definecolor{LightSalmon4}{rgb}{0.55,0.34,0.26}
\definecolor{LightSalmon}{rgb}{1.00,0.63,0.48}
\definecolor{LightSeaGreen}{rgb}{0.13,0.70,0.67}
\definecolor{LightSkyBlue1}{rgb}{0.69,0.89,1.00}
\definecolor{LightSkyBlue2}{rgb}{0.64,0.83,0.93}
\definecolor{LightSkyBlue3}{rgb}{0.55,0.71,0.80}
\definecolor{LightSkyBlue4}{rgb}{0.38,0.48,0.55}
\definecolor{LightSkyBlue}{rgb}{0.53,0.81,0.98}
\definecolor{LightSlateBlue}{rgb}{0.52,0.44,1.00}
\definecolor{LightSlateGray}{rgb}{0.47,0.53,0.60}
\definecolor{LightSlateGrey}{rgb}{0.47,0.53,0.60}
\definecolor{LightSteelBlue1}{rgb}{0.79,0.88,1.00}
\definecolor{LightSteelBlue2}{rgb}{0.74,0.82,0.93}
\definecolor{LightSteelBlue3}{rgb}{0.64,0.71,0.80}
\definecolor{LightSteelBlue4}{rgb}{0.43,0.48,0.55}
\definecolor{LightSteelBlue}{rgb}{0.69,0.77,0.87}
\definecolor{LightYellow1}{rgb}{1.00,1.00,0.88}
\definecolor{LightYellow2}{rgb}{0.93,0.93,0.82}
\definecolor{LightYellow3}{rgb}{0.80,0.80,0.71}
\definecolor{LightYellow4}{rgb}{0.55,0.55,0.48}
\definecolor{LightYellow}{rgb}{1.00,1.00,0.88}
\definecolor{LimeGreen}{rgb}{0.20,0.80,0.20}
\definecolor{MediumAquamarine}{rgb}{0.40,0.80,0.67}
\definecolor{MediumBlue}{rgb}{0.00,0.00,0.80}
\definecolor{MediumOrchid1}{rgb}{0.88,0.40,1.00}
\definecolor{MediumOrchid2}{rgb}{0.82,0.37,0.93}
\definecolor{MediumOrchid3}{rgb}{0.71,0.32,0.80}
\definecolor{MediumOrchid4}{rgb}{0.48,0.22,0.55}
\definecolor{MediumOrchid}{rgb}{0.73,0.33,0.83}
\definecolor{MediumPurple1}{rgb}{0.67,0.51,1.00}
\definecolor{MediumPurple2}{rgb}{0.62,0.47,0.93}
\definecolor{MediumPurple3}{rgb}{0.54,0.41,0.80}
\definecolor{MediumPurple4}{rgb}{0.36,0.28,0.55}
\definecolor{MediumPurple}{rgb}{0.58,0.44,0.86}
\definecolor{MediumSeaGreen}{rgb}{0.24,0.70,0.44}
\definecolor{MediumSlateBlue}{rgb}{0.48,0.41,0.93}
\definecolor{MediumSpringGreen}{rgb}{0.00,0.98,0.60}
\definecolor{MediumTurquoise}{rgb}{0.28,0.82,0.80}
\definecolor{MediumVioletRed}{rgb}{0.78,0.08,0.52}
\definecolor{MidnightBlue}{rgb}{0.10,0.10,0.44}
\definecolor{MintCream}{rgb}{0.96,1.00,0.98}
\definecolor{MistyRose1}{rgb}{1.00,0.89,0.88}
\definecolor{MistyRose2}{rgb}{0.93,0.84,0.82}
\definecolor{MistyRose3}{rgb}{0.80,0.72,0.71}
\definecolor{MistyRose4}{rgb}{0.55,0.49,0.48}
\definecolor{MistyRose}{rgb}{1.00,0.89,0.88}
\definecolor{NavajoWhite1}{rgb}{1.00,0.87,0.68}
\definecolor{NavajoWhite2}{rgb}{0.93,0.81,0.63}
\definecolor{NavajoWhite3}{rgb}{0.80,0.70,0.55}
\definecolor{NavajoWhite4}{rgb}{0.55,0.47,0.37}
\definecolor{NavajoWhite}{rgb}{1.00,0.87,0.68}
\definecolor{NavyBlue}{rgb}{0.00,0.00,0.50}
\definecolor{OldLace}{rgb}{0.99,0.96,0.90}
\definecolor{OliveDrab1}{rgb}{0.75,1.00,0.24}
\definecolor{OliveDrab2}{rgb}{0.70,0.93,0.23}
\definecolor{OliveDrab3}{rgb}{0.60,0.80,0.20}
\definecolor{OliveDrab4}{rgb}{0.41,0.55,0.13}
\definecolor{OliveDrab}{rgb}{0.42,0.56,0.14}
\definecolor{OrangeRed1}{rgb}{1.00,0.27,0.00}
\definecolor{OrangeRed2}{rgb}{0.93,0.25,0.00}
\definecolor{OrangeRed3}{rgb}{0.80,0.22,0.00}
\definecolor{OrangeRed4}{rgb}{0.55,0.15,0.00}
\definecolor{OrangeRed}{rgb}{1.00,0.27,0.00}
\definecolor{PaleGoldenrod}{rgb}{0.93,0.91,0.67}
\definecolor{PaleGreen1}{rgb}{0.60,1.00,0.60}
\definecolor{PaleGreen2}{rgb}{0.56,0.93,0.56}
\definecolor{PaleGreen3}{rgb}{0.49,0.80,0.49}
\definecolor{PaleGreen4}{rgb}{0.33,0.55,0.33}
\definecolor{PaleGreen}{rgb}{0.60,0.98,0.60}
\definecolor{PaleTurquoise1}{rgb}{0.73,1.00,1.00}
\definecolor{PaleTurquoise2}{rgb}{0.68,0.93,0.93}
\definecolor{PaleTurquoise3}{rgb}{0.59,0.80,0.80}
\definecolor{PaleTurquoise4}{rgb}{0.40,0.55,0.55}
\definecolor{PaleTurquoise}{rgb}{0.69,0.93,0.93}
\definecolor{PaleVioletRed1}{rgb}{1.00,0.51,0.67}
\definecolor{PaleVioletRed2}{rgb}{0.93,0.47,0.62}
\definecolor{PaleVioletRed3}{rgb}{0.80,0.41,0.54}
\definecolor{PaleVioletRed4}{rgb}{0.55,0.28,0.36}
\definecolor{PaleVioletRed}{rgb}{0.86,0.44,0.58}
\definecolor{PapayaWhip}{rgb}{1.00,0.94,0.84}
\definecolor{PeachPuff1}{rgb}{1.00,0.85,0.73}
\definecolor{PeachPuff2}{rgb}{0.93,0.80,0.68}
\definecolor{PeachPuff3}{rgb}{0.80,0.69,0.58}
\definecolor{PeachPuff4}{rgb}{0.55,0.47,0.40}
\definecolor{PeachPuff}{rgb}{1.00,0.85,0.73}
\definecolor{PowderBlue}{rgb}{0.69,0.88,0.90}
\definecolor{RosyBrown1}{rgb}{1.00,0.76,0.76}
\definecolor{RosyBrown2}{rgb}{0.93,0.71,0.71}
\definecolor{RosyBrown3}{rgb}{0.80,0.61,0.61}
\definecolor{RosyBrown4}{rgb}{0.55,0.41,0.41}
\definecolor{RosyBrown}{rgb}{0.74,0.56,0.56}
\definecolor{RoyalBlue1}{rgb}{0.28,0.46,1.00}
\definecolor{RoyalBlue2}{rgb}{0.26,0.43,0.93}
\definecolor{RoyalBlue3}{rgb}{0.23,0.37,0.80}
\definecolor{RoyalBlue4}{rgb}{0.15,0.25,0.55}
\definecolor{RoyalBlue}{rgb}{0.25,0.41,0.88}
\definecolor{SaddleBrown}{rgb}{0.55,0.27,0.07}
\definecolor{SandyBrown}{rgb}{0.96,0.64,0.38}
\definecolor{SeaGreen1}{rgb}{0.33,1.00,0.62}
\definecolor{SeaGreen2}{rgb}{0.31,0.93,0.58}
\definecolor{SeaGreen3}{rgb}{0.26,0.80,0.50}
\definecolor{SeaGreen4}{rgb}{0.18,0.55,0.34}
\definecolor{SeaGreen}{rgb}{0.18,0.55,0.34}
\definecolor{SkyBlue1}{rgb}{0.53,0.81,1.00}
\definecolor{SkyBlue2}{rgb}{0.49,0.75,0.93}
\definecolor{SkyBlue3}{rgb}{0.42,0.65,0.80}
\definecolor{SkyBlue4}{rgb}{0.29,0.44,0.55}
\definecolor{SkyBlue}{rgb}{0.53,0.81,0.92}
\definecolor{SlateBlue1}{rgb}{0.51,0.44,1.00}
\definecolor{SlateBlue2}{rgb}{0.48,0.40,0.93}
\definecolor{SlateBlue3}{rgb}{0.41,0.35,0.80}
\definecolor{SlateBlue4}{rgb}{0.28,0.24,0.55}
\definecolor{SlateBlue}{rgb}{0.42,0.35,0.80}
\definecolor{SlateGray1}{rgb}{0.78,0.89,1.00}
\definecolor{SlateGray2}{rgb}{0.73,0.83,0.93}
\definecolor{SlateGray3}{rgb}{0.62,0.71,0.80}
\definecolor{SlateGray4}{rgb}{0.42,0.48,0.55}
\definecolor{SlateGray}{rgb}{0.44,0.50,0.56}
\definecolor{SlateGrey}{rgb}{0.44,0.50,0.56}
\definecolor{SpringGreen1}{rgb}{0.00,1.00,0.50}
\definecolor{SpringGreen2}{rgb}{0.00,0.93,0.46}
\definecolor{SpringGreen3}{rgb}{0.00,0.80,0.40}
\definecolor{SpringGreen4}{rgb}{0.00,0.55,0.27}
\definecolor{SpringGreen}{rgb}{0.00,1.00,0.50}
\definecolor{SteelBlue1}{rgb}{0.39,0.72,1.00}
\definecolor{SteelBlue2}{rgb}{0.36,0.67,0.93}
\definecolor{SteelBlue3}{rgb}{0.31,0.58,0.80}
\definecolor{SteelBlue4}{rgb}{0.21,0.39,0.55}
\definecolor{SteelBlue}{rgb}{0.27,0.51,0.71}
\definecolor{VioletRed1}{rgb}{1.00,0.24,0.59}
\definecolor{VioletRed2}{rgb}{0.93,0.23,0.55}
\definecolor{VioletRed3}{rgb}{0.80,0.20,0.47}
\definecolor{VioletRed4}{rgb}{0.55,0.13,0.32}
\definecolor{VioletRed}{rgb}{0.82,0.13,0.56}
\definecolor{WhiteSmoke}{rgb}{0.96,0.96,0.96}
\definecolor{YellowGreen}{rgb}{0.60,0.80,0.20}
\definecolor{aliceblue}{rgb}{0.94,0.97,1.00}
\definecolor{antiquewhite}{rgb}{0.98,0.92,0.84}
\definecolor{aquamarine1}{rgb}{0.50,1.00,0.83}
\definecolor{aquamarine2}{rgb}{0.46,0.93,0.78}
\definecolor{aquamarine3}{rgb}{0.40,0.80,0.67}
\definecolor{aquamarine4}{rgb}{0.27,0.55,0.45}
\definecolor{aquamarine}{rgb}{0.50,1.00,0.83}
\definecolor{azure1}{rgb}{0.94,1.00,1.00}
\definecolor{azure2}{rgb}{0.88,0.93,0.93}
\definecolor{azure3}{rgb}{0.76,0.80,0.80}
\definecolor{azure4}{rgb}{0.51,0.55,0.55}
\definecolor{azure}{rgb}{0.94,1.00,1.00}
\definecolor{beige}{rgb}{0.96,0.96,0.86}
\definecolor{bisque1}{rgb}{1.00,0.89,0.77}
\definecolor{bisque2}{rgb}{0.93,0.84,0.72}
\definecolor{bisque3}{rgb}{0.80,0.72,0.62}
\definecolor{bisque4}{rgb}{0.55,0.49,0.42}
\definecolor{bisque}{rgb}{1.00,0.89,0.77}
\definecolor{black}{rgb}{0.00,0.00,0.00}
\definecolor{blanchedalmond}{rgb}{1.00,0.92,0.80}
\definecolor{blue1}{rgb}{0.00,0.00,1.00}
\definecolor{blue2}{rgb}{0.00,0.00,0.93}
\definecolor{blue3}{rgb}{0.00,0.00,0.80}
\definecolor{blue4}{rgb}{0.00,0.00,0.55}
\definecolor{blueviolet}{rgb}{0.54,0.17,0.89}
\definecolor{blue}{rgb}{0.00,0.00,1.00}
\definecolor{brown1}{rgb}{1.00,0.25,0.25}
\definecolor{brown2}{rgb}{0.93,0.23,0.23}
\definecolor{brown3}{rgb}{0.80,0.20,0.20}
\definecolor{brown4}{rgb}{0.55,0.14,0.14}
\definecolor{brown}{rgb}{0.65,0.16,0.16}
\definecolor{burlywood1}{rgb}{1.00,0.83,0.61}
\definecolor{burlywood2}{rgb}{0.93,0.77,0.57}
\definecolor{burlywood3}{rgb}{0.80,0.67,0.49}
\definecolor{burlywood4}{rgb}{0.55,0.45,0.33}
\definecolor{burlywood}{rgb}{0.87,0.72,0.53}
\definecolor{cadetblue}{rgb}{0.37,0.62,0.63}
\definecolor{chartreuse1}{rgb}{0.50,1.00,0.00}
\definecolor{chartreuse2}{rgb}{0.46,0.93,0.00}
\definecolor{chartreuse3}{rgb}{0.40,0.80,0.00}
\definecolor{chartreuse4}{rgb}{0.27,0.55,0.00}
\definecolor{chartreuse}{rgb}{0.50,1.00,0.00}
\definecolor{chocolate1}{rgb}{1.00,0.50,0.14}
\definecolor{chocolate2}{rgb}{0.93,0.46,0.13}
\definecolor{chocolate3}{rgb}{0.80,0.40,0.11}
\definecolor{chocolate4}{rgb}{0.55,0.27,0.07}
\definecolor{chocolate}{rgb}{0.82,0.41,0.12}
\definecolor{coral1}{rgb}{1.00,0.45,0.34}
\definecolor{coral2}{rgb}{0.93,0.42,0.31}
\definecolor{coral3}{rgb}{0.80,0.36,0.27}
\definecolor{coral4}{rgb}{0.55,0.24,0.18}
\definecolor{coral}{rgb}{1.00,0.50,0.31}
\definecolor{cornflowerblue}{rgb}{0.39,0.58,0.93}
\definecolor{cornsilk1}{rgb}{1.00,0.97,0.86}
\definecolor{cornsilk2}{rgb}{0.93,0.91,0.80}
\definecolor{cornsilk3}{rgb}{0.80,0.78,0.69}
\definecolor{cornsilk4}{rgb}{0.55,0.53,0.47}
\definecolor{cornsilk}{rgb}{1.00,0.97,0.86}
\definecolor{cyan1}{rgb}{0.00,1.00,1.00}
\definecolor{cyan2}{rgb}{0.00,0.93,0.93}
\definecolor{cyan3}{rgb}{0.00,0.80,0.80}
\definecolor{cyan4}{rgb}{0.00,0.55,0.55}
\definecolor{cyan}{rgb}{0.00,1.00,1.00}
\definecolor{darkblue}{rgb}{0.00,0.00,0.55}
\definecolor{darkcyan}{rgb}{0.00,0.55,0.55}
\definecolor{darkgoldenrod}{rgb}{0.72,0.53,0.04}
\definecolor{darkgray}{rgb}{0.66,0.66,0.66}
\definecolor{darkgreen}{rgb}{0.00,0.39,0.00}
\definecolor{darkgrey}{rgb}{0.66,0.66,0.66}
\definecolor{darkkhaki}{rgb}{0.74,0.72,0.42}
\definecolor{darkmagenta}{rgb}{0.55,0.00,0.55}
\definecolor{darkolive}{rgb}{0.33,0.42,0.18}
\definecolor{darkorange}{rgb}{1.00,0.55,0.00}
\definecolor{darkorchid}{rgb}{0.60,0.20,0.80}
\definecolor{darkred}{rgb}{0.55,0.00,0.00}
\definecolor{darksalmon}{rgb}{0.91,0.59,0.48}
\definecolor{darksea}{rgb}{0.56,0.74,0.56}
\definecolor{darkslate}{rgb}{0.18,0.31,0.31}
\definecolor{darkslate}{rgb}{0.18,0.31,0.31}
\definecolor{darkslate}{rgb}{0.28,0.24,0.55}
\definecolor{darkturquoise}{rgb}{0.00,0.81,0.82}
\definecolor{darkviolet}{rgb}{0.58,0.00,0.83}
\definecolor{deeppink}{rgb}{1.00,0.08,0.58}
\definecolor{deepsky}{rgb}{0.00,0.75,1.00}
\definecolor{dimgray}{rgb}{0.41,0.41,0.41}
\definecolor{dimgrey}{rgb}{0.41,0.41,0.41}
\definecolor{dodgerblue}{rgb}{0.12,0.56,1.00}
\definecolor{firebrick1}{rgb}{1.00,0.19,0.19}
\definecolor{firebrick2}{rgb}{0.93,0.17,0.17}
\definecolor{firebrick3}{rgb}{0.80,0.15,0.15}
\definecolor{firebrick4}{rgb}{0.55,0.10,0.10}
\definecolor{firebrick}{rgb}{0.70,0.13,0.13}
\definecolor{floralwhite}{rgb}{1.00,0.98,0.94}
\definecolor{forestgreen}{rgb}{0.13,0.55,0.13}
\definecolor{gainsboro}{rgb}{0.86,0.86,0.86}
\definecolor{ghostwhite}{rgb}{0.97,0.97,1.00}
\definecolor{gold1}{rgb}{1.00,0.84,0.00}
\definecolor{gold2}{rgb}{0.93,0.79,0.00}
\definecolor{gold3}{rgb}{0.80,0.68,0.00}
\definecolor{gold4}{rgb}{0.55,0.46,0.00}
\definecolor{goldenrod1}{rgb}{1.00,0.76,0.15}
\definecolor{goldenrod2}{rgb}{0.93,0.71,0.13}
\definecolor{goldenrod3}{rgb}{0.80,0.61,0.11}
\definecolor{goldenrod4}{rgb}{0.55,0.41,0.08}
\definecolor{goldenrod}{rgb}{0.85,0.65,0.13}
\definecolor{gold}{rgb}{1.00,0.84,0.00}
\definecolor{gray0}{rgb}{0.00,0.00,0.00}
\definecolor{gray100}{rgb}{1.00,1.00,1.00}
\definecolor{gray10}{rgb}{0.10,0.10,0.10}
\definecolor{gray11}{rgb}{0.11,0.11,0.11}
\definecolor{gray12}{rgb}{0.12,0.12,0.12}
\definecolor{gray13}{rgb}{0.13,0.13,0.13}
\definecolor{gray14}{rgb}{0.14,0.14,0.14}
\definecolor{gray15}{rgb}{0.15,0.15,0.15}
\definecolor{gray16}{rgb}{0.16,0.16,0.16}
\definecolor{gray17}{rgb}{0.17,0.17,0.17}
\definecolor{gray18}{rgb}{0.18,0.18,0.18}
\definecolor{gray19}{rgb}{0.19,0.19,0.19}
\definecolor{gray1}{rgb}{0.01,0.01,0.01}
\definecolor{gray20}{rgb}{0.20,0.20,0.20}
\definecolor{gray21}{rgb}{0.21,0.21,0.21}
\definecolor{gray22}{rgb}{0.22,0.22,0.22}
\definecolor{gray23}{rgb}{0.23,0.23,0.23}
\definecolor{gray24}{rgb}{0.24,0.24,0.24}
\definecolor{gray25}{rgb}{0.25,0.25,0.25}
\definecolor{gray26}{rgb}{0.26,0.26,0.26}
\definecolor{gray27}{rgb}{0.27,0.27,0.27}
\definecolor{gray28}{rgb}{0.28,0.28,0.28}
\definecolor{gray29}{rgb}{0.29,0.29,0.29}
\definecolor{gray2}{rgb}{0.02,0.02,0.02}
\definecolor{gray30}{rgb}{0.30,0.30,0.30}
\definecolor{gray31}{rgb}{0.31,0.31,0.31}
\definecolor{gray32}{rgb}{0.32,0.32,0.32}
\definecolor{gray33}{rgb}{0.33,0.33,0.33}
\definecolor{gray34}{rgb}{0.34,0.34,0.34}
\definecolor{gray35}{rgb}{0.35,0.35,0.35}
\definecolor{gray36}{rgb}{0.36,0.36,0.36}
\definecolor{gray37}{rgb}{0.37,0.37,0.37}
\definecolor{gray38}{rgb}{0.38,0.38,0.38}
\definecolor{gray39}{rgb}{0.39,0.39,0.39}
\definecolor{gray3}{rgb}{0.03,0.03,0.03}
\definecolor{gray40}{rgb}{0.40,0.40,0.40}
\definecolor{gray41}{rgb}{0.41,0.41,0.41}
\definecolor{gray42}{rgb}{0.42,0.42,0.42}
\definecolor{gray43}{rgb}{0.43,0.43,0.43}
\definecolor{gray44}{rgb}{0.44,0.44,0.44}
\definecolor{gray45}{rgb}{0.45,0.45,0.45}
\definecolor{gray46}{rgb}{0.46,0.46,0.46}
\definecolor{gray47}{rgb}{0.47,0.47,0.47}
\definecolor{gray48}{rgb}{0.48,0.48,0.48}
\definecolor{gray49}{rgb}{0.49,0.49,0.49}
\definecolor{gray4}{rgb}{0.04,0.04,0.04}
\definecolor{gray50}{rgb}{0.50,0.50,0.50}
\definecolor{gray51}{rgb}{0.51,0.51,0.51}
\definecolor{gray52}{rgb}{0.52,0.52,0.52}
\definecolor{gray53}{rgb}{0.53,0.53,0.53}
\definecolor{gray54}{rgb}{0.54,0.54,0.54}
\definecolor{gray55}{rgb}{0.55,0.55,0.55}
\definecolor{gray56}{rgb}{0.56,0.56,0.56}
\definecolor{gray57}{rgb}{0.57,0.57,0.57}
\definecolor{gray58}{rgb}{0.58,0.58,0.58}
\definecolor{gray59}{rgb}{0.59,0.59,0.59}
\definecolor{gray5}{rgb}{0.05,0.05,0.05}
\definecolor{gray60}{rgb}{0.60,0.60,0.60}
\definecolor{gray61}{rgb}{0.61,0.61,0.61}
\definecolor{gray62}{rgb}{0.62,0.62,0.62}
\definecolor{gray63}{rgb}{0.63,0.63,0.63}
\definecolor{gray64}{rgb}{0.64,0.64,0.64}
\definecolor{gray65}{rgb}{0.65,0.65,0.65}
\definecolor{gray66}{rgb}{0.66,0.66,0.66}
\definecolor{gray67}{rgb}{0.67,0.67,0.67}
\definecolor{gray68}{rgb}{0.68,0.68,0.68}
\definecolor{gray69}{rgb}{0.69,0.69,0.69}
\definecolor{gray6}{rgb}{0.06,0.06,0.06}
\definecolor{gray70}{rgb}{0.70,0.70,0.70}
\definecolor{gray71}{rgb}{0.71,0.71,0.71}
\definecolor{gray72}{rgb}{0.72,0.72,0.72}
\definecolor{gray73}{rgb}{0.73,0.73,0.73}
\definecolor{gray74}{rgb}{0.74,0.74,0.74}
\definecolor{gray75}{rgb}{0.75,0.75,0.75}
\definecolor{gray76}{rgb}{0.76,0.76,0.76}
\definecolor{gray77}{rgb}{0.77,0.77,0.77}
\definecolor{gray78}{rgb}{0.78,0.78,0.78}
\definecolor{gray79}{rgb}{0.79,0.79,0.79}
\definecolor{gray7}{rgb}{0.07,0.07,0.07}
\definecolor{gray80}{rgb}{0.80,0.80,0.80}
\definecolor{gray81}{rgb}{0.81,0.81,0.81}
\definecolor{gray82}{rgb}{0.82,0.82,0.82}
\definecolor{gray83}{rgb}{0.83,0.83,0.83}
\definecolor{gray84}{rgb}{0.84,0.84,0.84}
\definecolor{gray85}{rgb}{0.85,0.85,0.85}
\definecolor{gray86}{rgb}{0.86,0.86,0.86}
\definecolor{gray87}{rgb}{0.87,0.87,0.87}
\definecolor{gray88}{rgb}{0.88,0.88,0.88}
\definecolor{gray89}{rgb}{0.89,0.89,0.89}
\definecolor{gray8}{rgb}{0.08,0.08,0.08}
\definecolor{gray90}{rgb}{0.90,0.90,0.90}
\definecolor{gray91}{rgb}{0.91,0.91,0.91}
\definecolor{gray92}{rgb}{0.92,0.92,0.92}
\definecolor{gray93}{rgb}{0.93,0.93,0.93}
\definecolor{gray94}{rgb}{0.94,0.94,0.94}
\definecolor{gray95}{rgb}{0.95,0.95,0.95}
\definecolor{gray96}{rgb}{0.96,0.96,0.96}
\definecolor{gray97}{rgb}{0.97,0.97,0.97}
\definecolor{gray98}{rgb}{0.98,0.98,0.98}
\definecolor{gray99}{rgb}{0.99,0.99,0.99}
\definecolor{gray9}{rgb}{0.09,0.09,0.09}
\definecolor{gray}{rgb}{0.75,0.75,0.75}
\definecolor{green1}{rgb}{0.00,1.00,0.00}
\definecolor{green2}{rgb}{0.00,0.93,0.00}
\definecolor{green3}{rgb}{0.00,0.80,0.00}
\definecolor{green4}{rgb}{0.00,0.55,0.00}
\definecolor{greenyellow}{rgb}{0.68,1.00,0.18}
\definecolor{green}{rgb}{0.00,1.00,0.00}
\definecolor{grey0}{rgb}{0.00,0.00,0.00}
\definecolor{grey100}{rgb}{1.00,1.00,1.00}
\definecolor{grey10}{rgb}{0.10,0.10,0.10}
\definecolor{grey11}{rgb}{0.11,0.11,0.11}
\definecolor{grey12}{rgb}{0.12,0.12,0.12}
\definecolor{grey13}{rgb}{0.13,0.13,0.13}
\definecolor{grey14}{rgb}{0.14,0.14,0.14}
\definecolor{grey15}{rgb}{0.15,0.15,0.15}
\definecolor{grey16}{rgb}{0.16,0.16,0.16}
\definecolor{grey17}{rgb}{0.17,0.17,0.17}
\definecolor{grey18}{rgb}{0.18,0.18,0.18}
\definecolor{grey19}{rgb}{0.19,0.19,0.19}
\definecolor{grey1}{rgb}{0.01,0.01,0.01}
\definecolor{grey20}{rgb}{0.20,0.20,0.20}
\definecolor{grey21}{rgb}{0.21,0.21,0.21}
\definecolor{grey22}{rgb}{0.22,0.22,0.22}
\definecolor{grey23}{rgb}{0.23,0.23,0.23}
\definecolor{grey24}{rgb}{0.24,0.24,0.24}
\definecolor{grey25}{rgb}{0.25,0.25,0.25}
\definecolor{grey26}{rgb}{0.26,0.26,0.26}
\definecolor{grey27}{rgb}{0.27,0.27,0.27}
\definecolor{grey28}{rgb}{0.28,0.28,0.28}
\definecolor{grey29}{rgb}{0.29,0.29,0.29}
\definecolor{grey2}{rgb}{0.02,0.02,0.02}
\definecolor{grey30}{rgb}{0.30,0.30,0.30}
\definecolor{grey31}{rgb}{0.31,0.31,0.31}
\definecolor{grey32}{rgb}{0.32,0.32,0.32}
\definecolor{grey33}{rgb}{0.33,0.33,0.33}
\definecolor{grey34}{rgb}{0.34,0.34,0.34}
\definecolor{grey35}{rgb}{0.35,0.35,0.35}
\definecolor{grey36}{rgb}{0.36,0.36,0.36}
\definecolor{grey37}{rgb}{0.37,0.37,0.37}
\definecolor{grey38}{rgb}{0.38,0.38,0.38}
\definecolor{grey39}{rgb}{0.39,0.39,0.39}
\definecolor{grey3}{rgb}{0.03,0.03,0.03}
\definecolor{grey40}{rgb}{0.40,0.40,0.40}
\definecolor{grey41}{rgb}{0.41,0.41,0.41}
\definecolor{grey42}{rgb}{0.42,0.42,0.42}
\definecolor{grey43}{rgb}{0.43,0.43,0.43}
\definecolor{grey44}{rgb}{0.44,0.44,0.44}
\definecolor{grey45}{rgb}{0.45,0.45,0.45}
\definecolor{grey46}{rgb}{0.46,0.46,0.46}
\definecolor{grey47}{rgb}{0.47,0.47,0.47}
\definecolor{grey48}{rgb}{0.48,0.48,0.48}
\definecolor{grey49}{rgb}{0.49,0.49,0.49}
\definecolor{grey4}{rgb}{0.04,0.04,0.04}
\definecolor{grey50}{rgb}{0.50,0.50,0.50}
\definecolor{grey51}{rgb}{0.51,0.51,0.51}
\definecolor{grey52}{rgb}{0.52,0.52,0.52}
\definecolor{grey53}{rgb}{0.53,0.53,0.53}
\definecolor{grey54}{rgb}{0.54,0.54,0.54}
\definecolor{grey55}{rgb}{0.55,0.55,0.55}
\definecolor{grey56}{rgb}{0.56,0.56,0.56}
\definecolor{grey57}{rgb}{0.57,0.57,0.57}
\definecolor{grey58}{rgb}{0.58,0.58,0.58}
\definecolor{grey59}{rgb}{0.59,0.59,0.59}
\definecolor{grey5}{rgb}{0.05,0.05,0.05}
\definecolor{grey60}{rgb}{0.60,0.60,0.60}
\definecolor{grey61}{rgb}{0.61,0.61,0.61}
\definecolor{grey62}{rgb}{0.62,0.62,0.62}
\definecolor{grey63}{rgb}{0.63,0.63,0.63}
\definecolor{grey64}{rgb}{0.64,0.64,0.64}
\definecolor{grey65}{rgb}{0.65,0.65,0.65}
\definecolor{grey66}{rgb}{0.66,0.66,0.66}
\definecolor{grey67}{rgb}{0.67,0.67,0.67}
\definecolor{grey68}{rgb}{0.68,0.68,0.68}
\definecolor{grey69}{rgb}{0.69,0.69,0.69}
\definecolor{grey6}{rgb}{0.06,0.06,0.06}
\definecolor{grey70}{rgb}{0.70,0.70,0.70}
\definecolor{grey71}{rgb}{0.71,0.71,0.71}
\definecolor{grey72}{rgb}{0.72,0.72,0.72}
\definecolor{grey73}{rgb}{0.73,0.73,0.73}
\definecolor{grey74}{rgb}{0.74,0.74,0.74}
\definecolor{grey75}{rgb}{0.75,0.75,0.75}
\definecolor{grey76}{rgb}{0.76,0.76,0.76}
\definecolor{grey77}{rgb}{0.77,0.77,0.77}
\definecolor{grey78}{rgb}{0.78,0.78,0.78}
\definecolor{grey79}{rgb}{0.79,0.79,0.79}
\definecolor{grey7}{rgb}{0.07,0.07,0.07}
\definecolor{grey80}{rgb}{0.80,0.80,0.80}
\definecolor{grey81}{rgb}{0.81,0.81,0.81}
\definecolor{grey82}{rgb}{0.82,0.82,0.82}
\definecolor{grey83}{rgb}{0.83,0.83,0.83}
\definecolor{grey84}{rgb}{0.84,0.84,0.84}
\definecolor{grey85}{rgb}{0.85,0.85,0.85}
\definecolor{grey86}{rgb}{0.86,0.86,0.86}
\definecolor{grey87}{rgb}{0.87,0.87,0.87}
\definecolor{grey88}{rgb}{0.88,0.88,0.88}
\definecolor{grey89}{rgb}{0.89,0.89,0.89}
\definecolor{grey8}{rgb}{0.08,0.08,0.08}
\definecolor{grey90}{rgb}{0.90,0.90,0.90}
\definecolor{grey91}{rgb}{0.91,0.91,0.91}
\definecolor{grey92}{rgb}{0.92,0.92,0.92}
\definecolor{grey93}{rgb}{0.93,0.93,0.93}
\definecolor{grey94}{rgb}{0.94,0.94,0.94}
\definecolor{grey95}{rgb}{0.95,0.95,0.95}
\definecolor{grey96}{rgb}{0.96,0.96,0.96}
\definecolor{grey97}{rgb}{0.97,0.97,0.97}
\definecolor{grey98}{rgb}{0.98,0.98,0.98}
\definecolor{grey99}{rgb}{0.99,0.99,0.99}
\definecolor{grey9}{rgb}{0.09,0.09,0.09}
\definecolor{grey}{rgb}{0.75,0.75,0.75}
\definecolor{honeydew1}{rgb}{0.94,1.00,0.94}
\definecolor{honeydew2}{rgb}{0.88,0.93,0.88}
\definecolor{honeydew3}{rgb}{0.76,0.80,0.76}
\definecolor{honeydew4}{rgb}{0.51,0.55,0.51}
\definecolor{honeydew}{rgb}{0.94,1.00,0.94}
\definecolor{hotpink}{rgb}{1.00,0.41,0.71}
\definecolor{indianred}{rgb}{0.80,0.36,0.36}
\definecolor{ivory1}{rgb}{1.00,1.00,0.94}
\definecolor{ivory2}{rgb}{0.93,0.93,0.88}
\definecolor{ivory3}{rgb}{0.80,0.80,0.76}
\definecolor{ivory4}{rgb}{0.55,0.55,0.51}
\definecolor{ivory}{rgb}{1.00,1.00,0.94}
\definecolor{khaki1}{rgb}{1.00,0.96,0.56}
\definecolor{khaki2}{rgb}{0.93,0.90,0.52}
\definecolor{khaki3}{rgb}{0.80,0.78,0.45}
\definecolor{khaki4}{rgb}{0.55,0.53,0.31}
\definecolor{khaki}{rgb}{0.94,0.90,0.55}
\definecolor{lavenderblush}{rgb}{1.00,0.94,0.96}
\definecolor{lavender}{rgb}{0.90,0.90,0.98}
\definecolor{lawngreen}{rgb}{0.49,0.99,0.00}
\definecolor{lemonchiffon}{rgb}{1.00,0.98,0.80}
\definecolor{lightblue}{rgb}{0.68,0.85,0.90}
\definecolor{lightcoral}{rgb}{0.94,0.50,0.50}
\definecolor{lightcyan}{rgb}{0.88,1.00,1.00}
\definecolor{lightgoldenrod}{rgb}{0.93,0.87,0.51}
\definecolor{lightgoldenrod}{rgb}{0.98,0.98,0.82}
\definecolor{lightgray}{rgb}{0.83,0.83,0.83}
\definecolor{lightgreen}{rgb}{0.56,0.93,0.56}
\definecolor{lightgrey}{rgb}{0.83,0.83,0.83}
\definecolor{lightpink}{rgb}{1.00,0.71,0.76}
\definecolor{lightsalmon}{rgb}{1.00,0.63,0.48}
\definecolor{lightsea}{rgb}{0.13,0.70,0.67}
\definecolor{lightsky}{rgb}{0.53,0.81,0.98}
\definecolor{lightslate}{rgb}{0.47,0.53,0.60}
\definecolor{lightslate}{rgb}{0.47,0.53,0.60}
\definecolor{lightslate}{rgb}{0.52,0.44,1.00}
\definecolor{lightsteel}{rgb}{0.69,0.77,0.87}
\definecolor{lightyellow}{rgb}{1.00,1.00,0.88}
\definecolor{limegreen}{rgb}{0.20,0.80,0.20}
\definecolor{linen}{rgb}{0.98,0.94,0.90}
\definecolor{magenta1}{rgb}{1.00,0.00,1.00}
\definecolor{magenta2}{rgb}{0.93,0.00,0.93}
\definecolor{magenta3}{rgb}{0.80,0.00,0.80}
\definecolor{magenta4}{rgb}{0.55,0.00,0.55}
\definecolor{magenta}{rgb}{1.00,0.00,1.00}
\definecolor{maroon1}{rgb}{1.00,0.20,0.70}
\definecolor{maroon2}{rgb}{0.93,0.19,0.65}
\definecolor{maroon3}{rgb}{0.80,0.16,0.56}
\definecolor{maroon4}{rgb}{0.55,0.11,0.38}
\definecolor{maroon}{rgb}{0.69,0.19,0.38}
\definecolor{mediumaquamarine}{rgb}{0.40,0.80,0.67}
\definecolor{mediumblue}{rgb}{0.00,0.00,0.80}
\definecolor{mediumorchid}{rgb}{0.73,0.33,0.83}
\definecolor{mediumpurple}{rgb}{0.58,0.44,0.86}
\definecolor{mediumsea}{rgb}{0.24,0.70,0.44}
\definecolor{mediumslate}{rgb}{0.48,0.41,0.93}
\definecolor{mediumspring}{rgb}{0.00,0.98,0.60}
\definecolor{mediumturquoise}{rgb}{0.28,0.82,0.80}
\definecolor{mediumviolet}{rgb}{0.78,0.08,0.52}
\definecolor{midnightblue}{rgb}{0.10,0.10,0.44}
\definecolor{mintcream}{rgb}{0.96,1.00,0.98}
\definecolor{mistyrose}{rgb}{1.00,0.89,0.88}
\definecolor{moccasin}{rgb}{1.00,0.89,0.71}
\definecolor{navajowhite}{rgb}{1.00,0.87,0.68}
\definecolor{navyblue}{rgb}{0.00,0.00,0.50}
\definecolor{navy}{rgb}{0.00,0.00,0.50}
\definecolor{oldlace}{rgb}{0.99,0.96,0.90}
\definecolor{olivedrab}{rgb}{0.42,0.56,0.14}
\definecolor{orange1}{rgb}{1.00,0.65,0.00}
\definecolor{orange2}{rgb}{0.93,0.60,0.00}
\definecolor{orange3}{rgb}{0.80,0.52,0.00}
\definecolor{orange4}{rgb}{0.55,0.35,0.00}
\definecolor{orangered}{rgb}{1.00,0.27,0.00}
\definecolor{orange}{rgb}{1.00,0.65,0.00}
\definecolor{orchid1}{rgb}{1.00,0.51,0.98}
\definecolor{orchid2}{rgb}{0.93,0.48,0.91}
\definecolor{orchid3}{rgb}{0.80,0.41,0.79}
\definecolor{orchid4}{rgb}{0.55,0.28,0.54}
\definecolor{orchid}{rgb}{0.85,0.44,0.84}
\definecolor{palegoldenrod}{rgb}{0.93,0.91,0.67}
\definecolor{palegreen}{rgb}{0.60,0.98,0.60}
\definecolor{paleturquoise}{rgb}{0.69,0.93,0.93}
\definecolor{paleviolet}{rgb}{0.86,0.44,0.58}
\definecolor{papayawhip}{rgb}{1.00,0.94,0.84}
\definecolor{peachpuff}{rgb}{1.00,0.85,0.73}
\definecolor{peru}{rgb}{0.80,0.52,0.25}
\definecolor{pink1}{rgb}{1.00,0.71,0.77}
\definecolor{pink2}{rgb}{0.93,0.66,0.72}
\definecolor{pink3}{rgb}{0.80,0.57,0.62}
\definecolor{pink4}{rgb}{0.55,0.39,0.42}
\definecolor{pink}{rgb}{1.00,0.75,0.80}
\definecolor{plum1}{rgb}{1.00,0.73,1.00}
\definecolor{plum2}{rgb}{0.93,0.68,0.93}
\definecolor{plum3}{rgb}{0.80,0.59,0.80}
\definecolor{plum4}{rgb}{0.55,0.40,0.55}
\definecolor{plum}{rgb}{0.87,0.63,0.87}
\definecolor{powderblue}{rgb}{0.69,0.88,0.90}
\definecolor{purple1}{rgb}{0.61,0.19,1.00}
\definecolor{purple2}{rgb}{0.57,0.17,0.93}
\definecolor{purple3}{rgb}{0.49,0.15,0.80}
\definecolor{purple4}{rgb}{0.33,0.10,0.55}
\definecolor{purple}{rgb}{0.63,0.13,0.94}
\definecolor{red1}{rgb}{1.00,0.00,0.00}
\definecolor{red2}{rgb}{0.93,0.00,0.00}
\definecolor{red3}{rgb}{0.80,0.00,0.00}
\definecolor{red4}{rgb}{0.55,0.00,0.00}
\definecolor{red}{rgb}{1.00,0.00,0.00}
\definecolor{rosybrown}{rgb}{0.74,0.56,0.56}
\definecolor{royalblue}{rgb}{0.25,0.41,0.88}
\definecolor{saddlebrown}{rgb}{0.55,0.27,0.07}
\definecolor{salmon1}{rgb}{1.00,0.55,0.41}
\definecolor{salmon2}{rgb}{0.93,0.51,0.38}
\definecolor{salmon3}{rgb}{0.80,0.44,0.33}
\definecolor{salmon4}{rgb}{0.55,0.30,0.22}
\definecolor{salmon}{rgb}{0.98,0.50,0.45}
\definecolor{sandybrown}{rgb}{0.96,0.64,0.38}
\definecolor{seagreen}{rgb}{0.18,0.55,0.34}
\definecolor{seashell1}{rgb}{1.00,0.96,0.93}
\definecolor{seashell2}{rgb}{0.93,0.90,0.87}
\definecolor{seashell3}{rgb}{0.80,0.77,0.75}
\definecolor{seashell4}{rgb}{0.55,0.53,0.51}
\definecolor{seashell}{rgb}{1.00,0.96,0.93}
\definecolor{sienna1}{rgb}{1.00,0.51,0.28}
\definecolor{sienna2}{rgb}{0.93,0.47,0.26}
\definecolor{sienna3}{rgb}{0.80,0.41,0.22}
\definecolor{sienna4}{rgb}{0.55,0.28,0.15}
\definecolor{sienna}{rgb}{0.63,0.32,0.18}
\definecolor{skyblue}{rgb}{0.53,0.81,0.92}
\definecolor{slateblue}{rgb}{0.42,0.35,0.80}
\definecolor{slategray}{rgb}{0.44,0.50,0.56}
\definecolor{slategrey}{rgb}{0.44,0.50,0.56}
\definecolor{snow1}{rgb}{1.00,0.98,0.98}
\definecolor{snow2}{rgb}{0.93,0.91,0.91}
\definecolor{snow3}{rgb}{0.80,0.79,0.79}
\definecolor{snow4}{rgb}{0.55,0.54,0.54}
\definecolor{snow}{rgb}{1.00,0.98,0.98}
\definecolor{springgreen}{rgb}{0.00,1.00,0.50}
\definecolor{steelblue}{rgb}{0.27,0.51,0.71}
\definecolor{tan1}{rgb}{1.00,0.65,0.31}
\definecolor{tan2}{rgb}{0.93,0.60,0.29}
\definecolor{tan3}{rgb}{0.80,0.52,0.25}
\definecolor{tan4}{rgb}{0.55,0.35,0.17}
\definecolor{tan}{rgb}{0.82,0.71,0.55}
\definecolor{thistle1}{rgb}{1.00,0.88,1.00}
\definecolor{thistle2}{rgb}{0.93,0.82,0.93}
\definecolor{thistle3}{rgb}{0.80,0.71,0.80}
\definecolor{thistle4}{rgb}{0.55,0.48,0.55}
\definecolor{thistle}{rgb}{0.85,0.75,0.85}
\definecolor{tomato1}{rgb}{1.00,0.39,0.28}
\definecolor{tomato2}{rgb}{0.93,0.36,0.26}
\definecolor{tomato3}{rgb}{0.80,0.31,0.22}
\definecolor{tomato4}{rgb}{0.55,0.21,0.15}
\definecolor{tomato}{rgb}{1.00,0.39,0.28}
\definecolor{turquoise1}{rgb}{0.00,0.96,1.00}
\definecolor{turquoise2}{rgb}{0.00,0.90,0.93}
\definecolor{turquoise3}{rgb}{0.00,0.77,0.80}
\definecolor{turquoise4}{rgb}{0.00,0.53,0.55}
\definecolor{turquoise}{rgb}{0.25,0.88,0.82}
\definecolor{violetred}{rgb}{0.82,0.13,0.56}
\definecolor{violet}{rgb}{0.93,0.51,0.93}
\definecolor{wheat1}{rgb}{1.00,0.91,0.73}
\definecolor{wheat2}{rgb}{0.93,0.85,0.68}
\definecolor{wheat3}{rgb}{0.80,0.73,0.59}
\definecolor{wheat4}{rgb}{0.55,0.49,0.40}
\definecolor{wheat}{rgb}{0.96,0.87,0.70}
\definecolor{whitesmoke}{rgb}{0.96,0.96,0.96}
\definecolor{white}{rgb}{1.00,1.00,1.00}
\definecolor{yellow1}{rgb}{1.00,1.00,0.00}
\definecolor{yellow2}{rgb}{0.93,0.93,0.00}
\definecolor{yellow3}{rgb}{0.80,0.80,0.00}
\definecolor{yellow4}{rgb}{0.55,0.55,0.00}
\definecolor{yellowgreen}{rgb}{0.60,0.80,0.20}
\definecolor{yellow}{rgb}{1.00,1.00,0.00}

\usepackage{graphicx}
\usepackage{color}
\usepackage{placeins}
\usepackage{float}
\usepackage{tabularx,colortbl}

\begin{document}
%
\title{A Parallel Compressive Imaging Architecture for One-Shot Acquisition}

\author{\IEEEauthorblockN{Tomas Bj\"{o}rklund, Enrico Magli}
\IEEEauthorblockA{Department of Electronics and Telecommunications\\
Politecnico di Torino, Turin, Italy\\}
}



\setlength{\textfloatsep}{5pt plus 1.0pt minus 2.0pt}
\setlength{\dbltextfloatsep}{0pt plus 1.0pt minus 2.0pt}
\setlength{\dblfloatsep }{0pt plus 1.0pt minus 2.0pt}
\setlength{\abovecaptionskip}{0pt} 

\addtolength{\abovedisplayskip}{-1ex}
\addtolength{\abovedisplayshortskip}{-1ex}
\addtolength{\belowdisplayskip}{-1ex}
\addtolength{\belowdisplayshortskip}{-1ex}

\maketitle

\begin{abstract}

A limitation of many compressive imaging architectures lies in the sequential nature of the sensing process, which leads to long sensing times. In this paper we present a novel architecture that uses fewer detectors than the number of reconstructed pixels and is able to acquire the image in a single acquisition. This paves the way for the development of video architectures that acquire several frames per second. We specifically address the diffraction problem, showing that deconvolution normally used to recover diffraction blur can be replaced by convolution of the sensing matrix, and how measurements of a 0/1 physical sensing matrix can be converted to -1/1 compressive sensing matrix without any extra acquisitions. Simulations of our architecture show that the image quality is comparable to that of a classic Compressive Imaging camera, whereas the proposed architecture avoids long acquisition times due to sequential sensing. This one-shot procedure also allows to employ a fixed sensing matrix instead of a complex device such as a Digital Micro Mirror array or Spatial Light Modulator. It also enables imaging at bandwidths where these are not efficient.

\end{abstract}



%
\IEEEpeerreviewmaketitle

\section{Introduction}
Compressed Sensing (CS) \cite{CS:Candes}\cite{CS:Donoho} is a novel framework for acquisition of compressible data at sub-Nyquist sampling rates, moving computational complexity from sensing phase to reconstruction. In Compressive Imaging (CI), CS is applied to reconstruct images from fewer measurements than the number of image pixels, under the condition that the image is sparse or at least compressible in some domain. The success of JPEG indicates that most natural images are highly compressible with only small losses of image quality. Seminal works on CI include the single-pixel camera \cite{CI:Duarte} and single-pixel terahertz imaging system \cite{CI:Chan}, which acquire the image through sequential measurements from a single sensor while changing random sensing patterns in front of it. Compressed coded aperture imaging \cite{CI:Marcia} uses a coded aperture to project overlapping coded copies of the image onto a detector array to obtain superresolution using CS. Similarly, CMOS compressive imagers \cite{CI:Jacques}\cite{CI:Oike} use detector arrays performing combinations of analog measurements before converting into fewer digital compressed measurements. This allows to significantly decrease power consumption.

A limitation of architectures based on the single-pixel camera lies in the sequential, and hence slow acquisition process. To some extent this can be addressed by block-based CS \cite{CI:blockbased} and {\color{black}\cite{CI:columnwise}}, which can in part parallelize the sensing process. In this paper we present a new CI framework which allows faster acquisition than \cite{CI:Duarte}\cite{CI:Chan}\cite{CI:blockbased}{\color{black}\cite{CI:columnwise}}, in which the total time is linearly proportional to the number of measurements. As in \cite{CI:Marcia} we use fewer detectors than \cite{CI:Jacques}\cite{CI:Oike}, but we also demonstrate that our architecture can be used even if diffraction from the sensing pattern is prominent; this enables smaller camera dimensions and the use of lower energy radiation for imaging. Besides shortening the total acquisition time, the proposed architecture also weakens the requirement on the modulation and acquisition rate of the sensing matrix and the detector array, allowing a cheaper and simpler construction, and paving the way for compressive video capture in real-time.

\section{Background: Compressed Sensing}
\noindent Consider the sensing process:
\begin{align}
y = A x \text{ ,}
\label{eqn:CSsensing}
\end{align}
where $x$ is the signal of interest, $y$ are the measurements and $A$ is an $r\times c$ sensing matrix with $r\ll c$. If $x$ is sparse in some domain and $A$  satisfies the Restricted Isometry Property (RIP) \cite{CI:CandesTao}, $x$ can be recovered with very high probability solving the minimization problem
\begin{align}
\min_{\tilde{x} \in \mathbb{R}^r} \left|\left|{\tilde{x}}\right|\right|_{1} \text{subject to  } y = A \tilde{x}  \text{ .}
\label{eqn:CSminimization}
\end{align}
For images, a common approach is to instead minimize the total variation norm \cite{CS:Li}
\vspace{0.1cm}
\begin{align*}
TV(x) = \sum_{i,j} \sqrt{\left|x_{i+1,j}-x_{i,j}\right|^2 + \left|x_{i,j+1}-x_{i,j}\right|^2}  \text{ ,}
\end{align*}

\noindent which assumes the image gradient to be sparse. This is the method used to recover images in this paper.

In CI, the sensing matrix can take the form of a physical filter. The filter modulates the light of each image pixel before it reaches the detector(s). Each measurement uses a different modulation pattern. The filter can be realized by a Digital Micromirror Device (DMD) as in \cite{CI:Duarte} or a Spatial Light Modulator (SLM) or plates with multiple fixed interchangeable filters with patterns of holes or transparencies \cite{CI:Chan}.



\section{Proposed Parallel Compressive Imaging Architecture}
The architecture we propose to parallelize the sensing process is illustrated in Fig. \ref{fig:OpticsOverview}. To simultaneously acquire measurements of multiple sensing patterns, the image is \emph{not focused} when projected onto the sensing matrix. The unfocused projection can be seen as \emph{shifted copies} of the image. {\color{black} These copies receive different encodings, which allows for parallel acquisition of measurements, without needing to update the sensing matrix between each single measurement}. Shifting a longer sensing pattern in one direction has been shown to work for CS reconstruction with little impact on the reconstruction \cite{CI:Heidari}. We will show later that the sensing matrix has a block-Toeplitz structure when shifting in two directions; this structure has been shown to satisfy the RIP property \cite{CI:Marcia}\cite{CS:Bajwa}. Fig. \ref{fig:OpticsOverview} shows an overview of the optical setup. After the target \emph{image}, a \emph{first lens} is located at distance $s_o$. This lens focuses an image at the distance $s_i$ such that $\frac{1}{s_o}+\frac{1}{s_i} = \frac{1}{f_1}$, where $f_1$ is the focal length. At distance $s_i$ from the lens, a diaphragm is placed with an \emph{aperture} for the focused image to prevent objects outside the target image region from interfering with the measurements. A two dimensional SLM is placed out of focus such that different sets of parallel beams of the image hit the modulator in a shifted and overlapping manner. Through the \emph{modulator} each set of parallel beams (i.e. each shifted image) receives a different encoding pattern. The distance from the aperture depends on the size and the diameter of the first lens such that all shifts are projected onto it. A second lens is positioned after the modulator to focus all parallel beams at one point in the focal plane. The position of the focal point for each set of parallel beams depends on the incident angle. Finally, a \emph{detector} array is positioned in the focal plane of the second lens. In this manner each pixel detects the focused light from a set of originally parallel beams, which corresponds to the sum of all image pixels, uniquely modulated by a specific shift of the pattern on the modulator. Moreover, Fig. \ref{fig:OpticsOverview} illustrates how a set of parallel beams at an angle $\alpha$ is modulated by the lowermost shift of the modulator and focused onto the lowest detector pixel, while the beamset parallel to the optical axis is modulated by the center shift and detected by the center pixel.

\subsection{Design of the Sensing Matrix}
\begin{figure}
\centering
\setlength{\abovecaptionskip}{0pt} 
\includegraphics[width=3.5in]{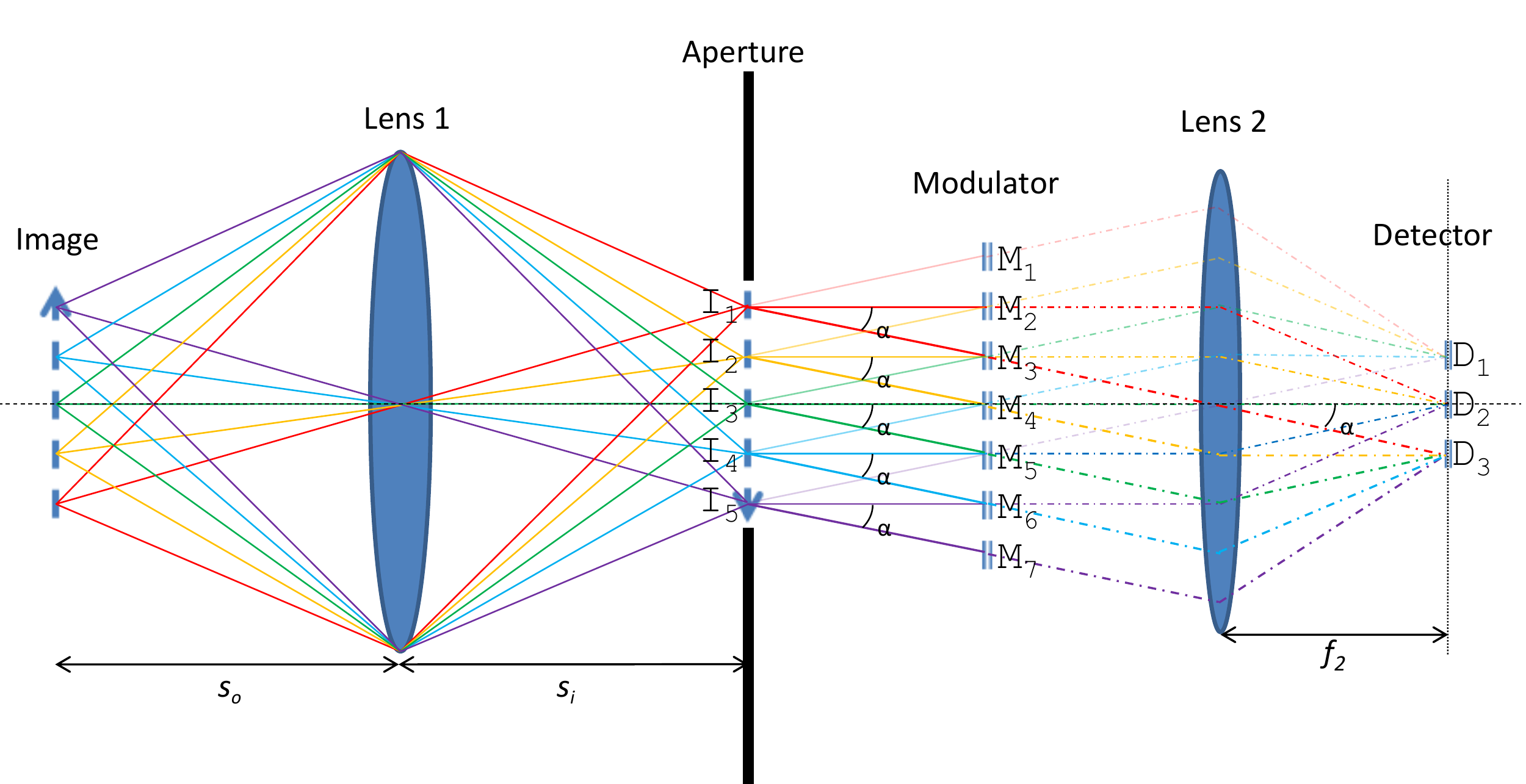}
\caption{Overview of the optics setup. The highlighted beams after the aperture illustrates how two sets of parallel beams at different incident angle at the modulator are modulated by shifted patterns. Each set of parallel beams is focused at a unique point at the detector.}
\label{fig:OpticsOverview}
\end{figure}

The light paths in the one-dimensional case are illustrated in Fig. \ref{fig:OpticsOverview}. The acquisition process is described by the following set of equations for 5 image points and 3 detectors:
\vspace{0.1cm}
\begin{align*}
\begin{cases} D_1 = M_1 I_1+M_2 I_2+M_3 I_3+M_4 I_4+M_5 I_5 \\  D_2 = M_2I_1+M_3I_2+M_4I_3+M_5I_4+M_6I_5 \\ D_3 = M_3I_1+M_4I_2+M_5I_3+M_6I_4+M_7I_5 \end{cases},
\end{align*}
\vspace{0.1cm}

\noindent where $I_i$ is the irradiance of the image point $i$, $M_j$ is the transmittance of modulator pixel $j$ and $D_k$ is the total irradiance at detector $k$. The equation system is linear and can be formulated as,
\vspace{-0.2cm}
\begin{align}
D=MI  \text{,}
\label{eqn:DMI}
\end{align}
\noindent
where $M$ is the sensing matrix, $I$ the image vector (not to be confused with the identity matrix) and $D$ the vector of detector measurements.

In a more realistic scenario the image, the modulator and the detector are all two-dimensional. To maintain a 2D form of the full sensing matrix, the image and sensor are row-wise wrapped into column vectors, and each measurement submatrix of the modulator matrix is column-wise wrapped into rows of the full sensing matrix. Let $\mathcal{D}$ be the $k\times l$ detector matrix and $\mathcal{I}$ the $m\times n$ image, and let $D=[\mathcal{D}_{1,1},  \hdots,   \mathcal{D}_{1,l},  \mathcal{D}_{2,1},  \hdots, \mathcal{D}_{k-1,l-1},  \mathcal{D}_{k,1}, \hdots, \mathcal{D}_{k,l} ]^T$  and $I=[\mathcal{I}_{1,1} , \hdots , \mathcal{I}_{m,1} , \mathcal{I}_{1,2}  , \hdots , \mathcal{I}_{m-1,n-1} ,  \mathcal{I}_{1,n} , \hdots  , \mathcal{I}_{m,n} ]^T$
be their vectorized forms. 

An undersampling of $\frac{1}{4}$ can be achieved by using matrix $\mathcal{D}$ of dimensions $k=\frac{m}{2}$ and $l=\frac{n}{2}$, this suggests that a modulator matrix $\mathcal{M}$ of size $(k-1+m)\times (l-1+n)$ is required to allow for $k$ vertical and $l$ horizontal shifts of $\mathcal{I}$. However, early experiments using this strategy were unsuccessful; we conjecture this is due to too high correlation between the measurements because neighboring pixels are too correlated in typical natural images. Instead we use $\mathcal{D}$ of dimensions $k=m$ and $l=n$ and $\mathcal{M}$ of size $(k+m)\times (l+n)$ and afterwards downsample $\mathcal{D}$ and $\mathcal{M}$ to fit the real detector dimensions, as described later. The sensing matrix $M$ of row-wise vectorized shifts then has the form:
\begin{gather}
\scalemath{0.535}{
M=\begin{bmatrix}
\mathcal{M}_{1,1}	& \mathcal{M}_{1,2} 	&  \cdots 	& \mathcal{M}_{1,n}	& \mathcal{M}_{2,1}	& \cdots 	& \mathcal{M}_{2,n}   	& \cdots\cdots 	& \mathcal{M}_{m,1}  	& \cdots 	& \mathcal{M}_{m,n}		\\
\mathcal{M}_{1,2}	&  \mathcal{M}_{1,3} 	&  \cdots 	& \mathcal{M}_{1,n+1}& \mathcal{M}_{2,2}	& \cdots 	& \mathcal{M}_{2,n+1}& \cdots\cdots 	& \mathcal{M}_{m,2}  	& \cdots 	& \mathcal{M}_{m,n+1}	\\
\vdots 	&  \vdots    	&  \ddots 	&\vdots  	& \vdots  	& \ddots 	& \vdots   	& \ddots 	&             	& \vdots   	& \ddots 	& \vdots  		\\
\mathcal{M}_{1,l}	& \mathcal{M}_{1,l+1}	&  \cdots 	& \mathcal{M}_{1,l+n}& \mathcal{M}_{2,l} 	& \cdots 	& \mathcal{M}_{2,l+n} & \cdots\cdots  	& \mathcal{M}_{m,l}  	& \cdots 	& \mathcal{M}_{m,l+n}	\\
\mathcal{M}_{2,1}	&  \mathcal{M}_{2,2} 	&  \cdots 	& \mathcal{M}_{2,n} 	& \mathcal{M}_{3,1} 	& \cdots 	& \mathcal{M}_{3,n}	& \cdots\cdots 	& \mathcal{M}_{m+1,1}& \cdots 	& \mathcal{M}_{m+1,n}	\\
\vdots 	&  \vdots    	&  \ddots 	&\vdots  	& \vdots    	& \ddots 	& \vdots   	& \ddots 	& \vdots   	& \ddots 	& \vdots  		\\
\mathcal{M}_{2,l}	&  \mathcal{M}_{2,l+1}&  \cdots 	& \mathcal{M}_{2,l+n}& \mathcal{M}_{3,l} 	& \cdots 	& \mathcal{M}_{3,l+n}& \cdots\cdots  	& \mathcal{M}_{m+1,l}& \cdots 	& \mathcal{M}_{m+1,n+l}	\\
\vspace{-0.25cm}
\vdots 	& \vdots    	&  \ddots 	&\vdots   	& \vdots    	& \ddots 	& \vdots    	& \ddots\phantom{\ddots} 	& \vdots  	& \ddots 	& \vdots  		\\
\vdots  	&  \vdots   	&             	& \vdots 	& \vdots    	&            	& \vdots   	& \phantom{\ddots}\ddots        	& \vdots  	&            	& \vdots  		\\
\mathcal{M}_{k,1} 	& \mathcal{M}_{k,2} 	&  \cdots & \mathcal{M}_{k,n}      	& \mathcal{M}_{k+1,1}	& \cdots 	& \mathcal{M}_{k+1,n}	& \cdots\cdots 	& \mathcal{M}_{k+m,1}& \cdots 	& \mathcal{M}_{k+m,n}	\\
\vdots   	& \vdots    	&  \ddots &\vdots         	& \vdots     	& \ddots 	& \vdots  	& \ddots  	& \vdots   	& \ddots 	& \vdots  		\\
\mathcal{M}_{k,l}	& \mathcal{M}_{k,l+1}	& \cdots & \mathcal{M}_{k,l+n} 	& \mathcal{M}_{k+1,l}	& \cdots 	& \mathcal{M}_{k+1,l+n}& \cdots\cdots 	& \mathcal{M}_{k+m,l}	& \cdots 	& \mathcal{M}_{k+m,l+n}	\\
 \end{bmatrix} 
\label{eqn:sensingmatrix}
}
\end{gather}

\noindent After the vectorization of $\mathcal{D}$, $\mathcal{I}$ and $\mathcal{M}$ the acquisition process can again be described by the matrix multiplication (\ref{eqn:DMI}). The size of the full sensing matrix $M$ is $(k\cdotp l)\times (m\cdotp n)$. Note that if $k=m$ and $l=n$ the matrix has a (left-shifted) blockwise Toeplitz form.

We will consider two different approaches based on undersampled acquisition using $\frac{m \cdotp n}{4}$ measurements based on (\ref{eqn:sensingmatrix}) for $k=m$ and $l=n$: 
\begin{itemize}
\item[A.] Double horizontal and vertical shifting by discarding measurements $\mathcal{D}_{i,j}$ where at least one of $i$ or $j$ is even \end{itemize}\vspace{0.0cm} (and likewise for each row in $M$ based on the first element $\mathcal{M}_{i,j}$, this is the same downsampling used in \cite{CI:Marcia}). This corresponds to a detector array with a fill factor of $\leq 25\%$ such that the discarded measurements are projected on the dead-space between the real detector units.
\begin{itemize}
\item[B.] Group all measurements $D_{i,j} + D_{i+1,j} + D_{i,j+1} + D_{i+1,j+1}$, where $i$ and $j$ are odd. This corresponds to \end{itemize} a detector array with half the amount of pixels both vertically and horizontally, in which every pixel is twice as large.

\subsection{Conversion of physical measurements to CS}
We limit the scope of this paper to a random sensing matrix with elements to take on value $-1/1$ with probability 0.5. In the physical matrix we use $0$ and $100\%$ transmittance respectively to represent $-1$ and $1$ since a negative transmittance is not possible. This requires a mathematical correction of the measurements by $D=2 D_{raw}-I_{total}$, where $D_{raw}$ is the vector of raw detector measurements and $I_{total}$ is the total irradiance of all image pixels without modulation ($100\%$ transmittance). $I_{total}$ can be determined using an extra acquisition with all sensing pixels open, as in \cite{CI:Duarte}. However, by constructing $\mathcal{M}$ such that $\mathcal{M}_{i,j}=\mathcal{M}_{m+i,j}=\mathcal{M}_{i,n+j}=\mathcal{M}_{m+i,n+j}$ with $1\leq i \leq m$, $1\leq j \leq n$, $k=m$ and $l=n$, each image pixel is sensed equally over all measurements because either $\mathcal{M}_{i,j}$ or one of the repeated twins is shifted over each image pixel exactly once. Then $I_{total}$ can be calculated as
\vspace{-0.2cm}
\begin{align*}
I_{total}=\frac{\sum_{i=1}^k \sum_{j=1}^l \mathcal{D}_{i,j}}{\sum_{i=1}^m \sum_{j=1}^n \mathcal{M}_{ i,j} }.
\end{align*}
\noindent With this limitation $M$ has the same block-Toeplitz form as \cite{CI:Marcia} which satisfies the RIP. Since architecture A discards measurements  in $D$  we still need a second acquisition with all sensing pixels open to measure $I_{total}$, but since architecture B only sums the measurements of $D$, this method can still be applied.

\vspace{0.0cm}
In this manner we can acquire all CS measurements in parallel and hence achieve a decrease of the acquisition time up to $\frac{1}{N}$, where $N$ is the number of measurements,  compared to a sequential acquisition process. Acquiring all measurements simultaneously also enables the use of a fixed sensing matrix {\color{black}(e.g. an opaque membrane with holes)}, this does not only simplify the construction but also significantly reduces the random numbers to be stored or generated for the sensing matrix from $N \cdotp m  \cdotp n$ down to $4 \cdotp m\cdotp n$. However, this comes at a price; by not focusing the image at the sensing matrix nor at the detector, diffraction at the apertures of the sensing matrix need to be considered.

\subsection{Diffraction Compensation of the Sensing Matrix}
\begin{figure*}
\centering
\subfigure[diffraction-free scenario]{
   \includegraphics[width=3.3cm] {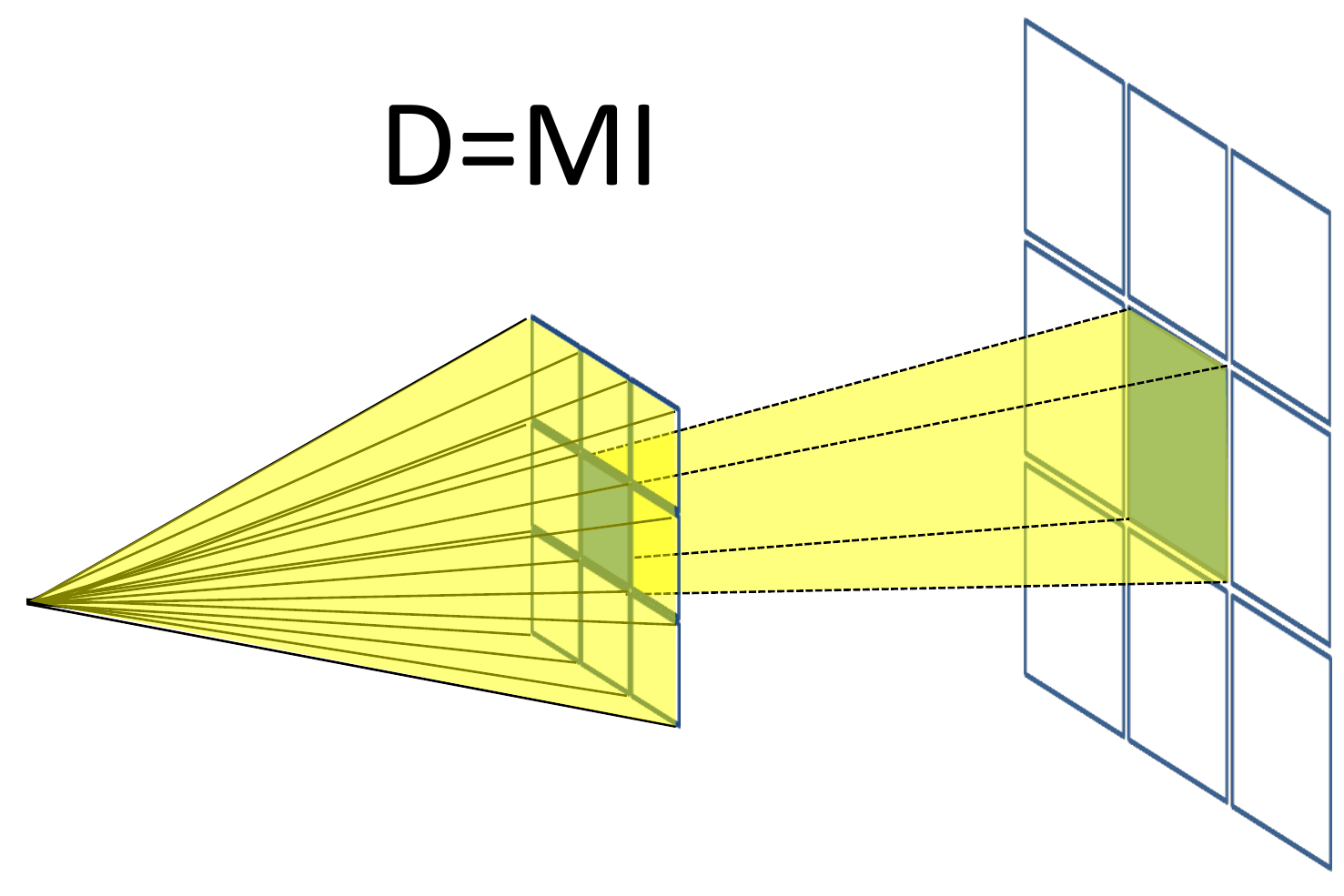}
   \label{fig:diffraction1}
 }
 \subfigure[diffraction]{
   \includegraphics[width=3.3cm] {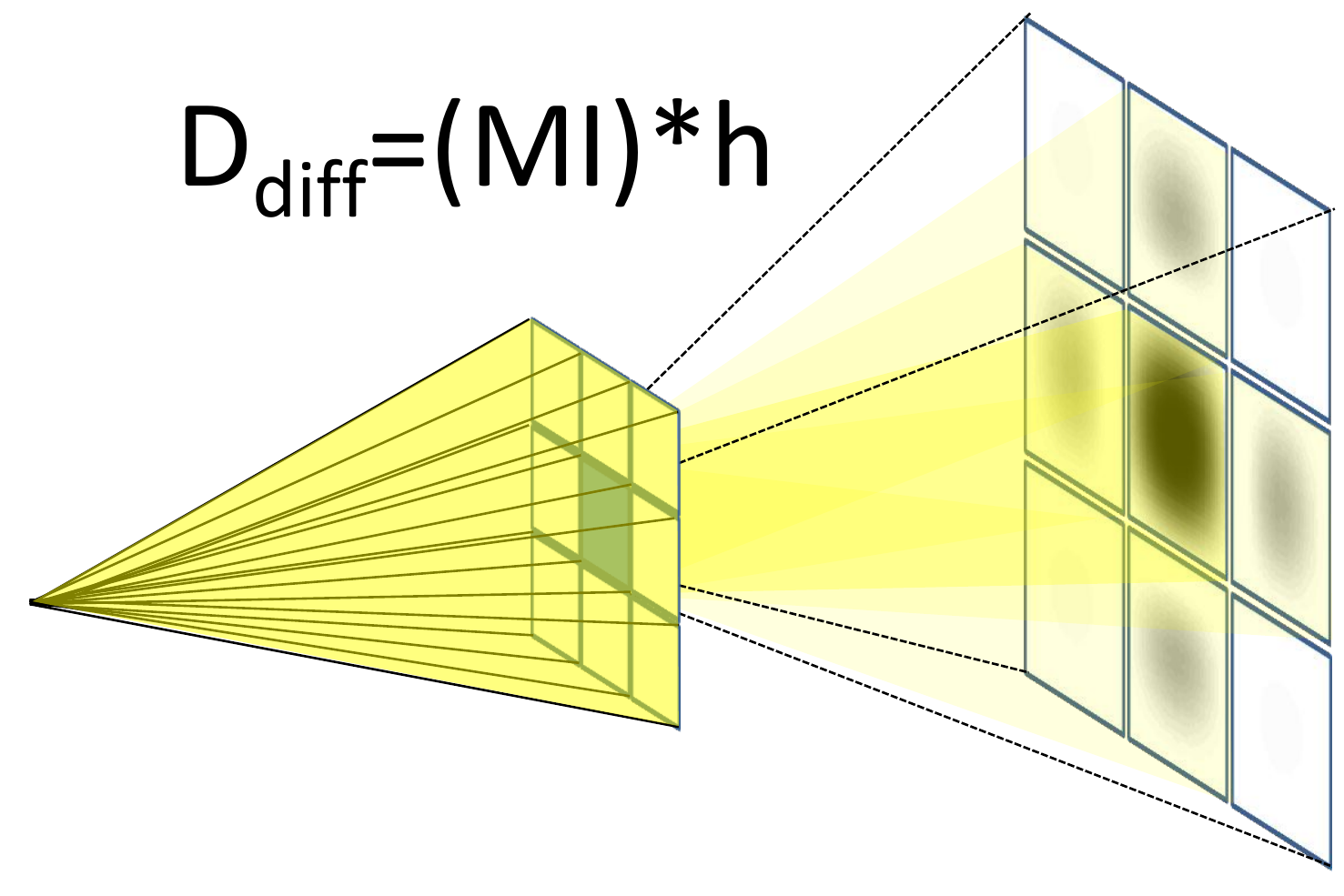}
   \label{fig:diffraction2}
 }
 \subfigure[diffraction model]{
   \includegraphics[width=3.3cm] {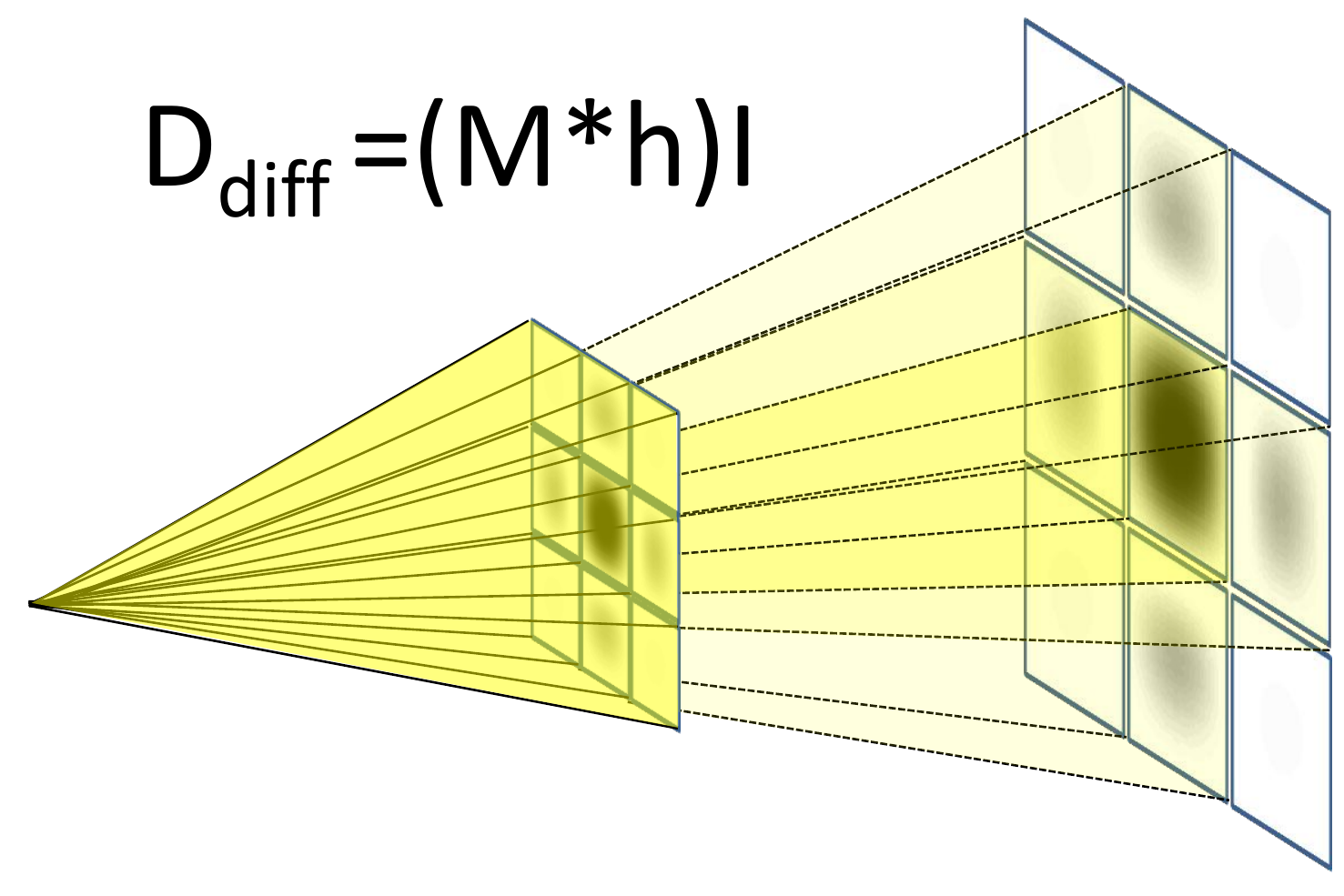}
   \label{fig:diffraction3}
 }
 \subfigure[PSF]{
   \includegraphics[width=3cm] {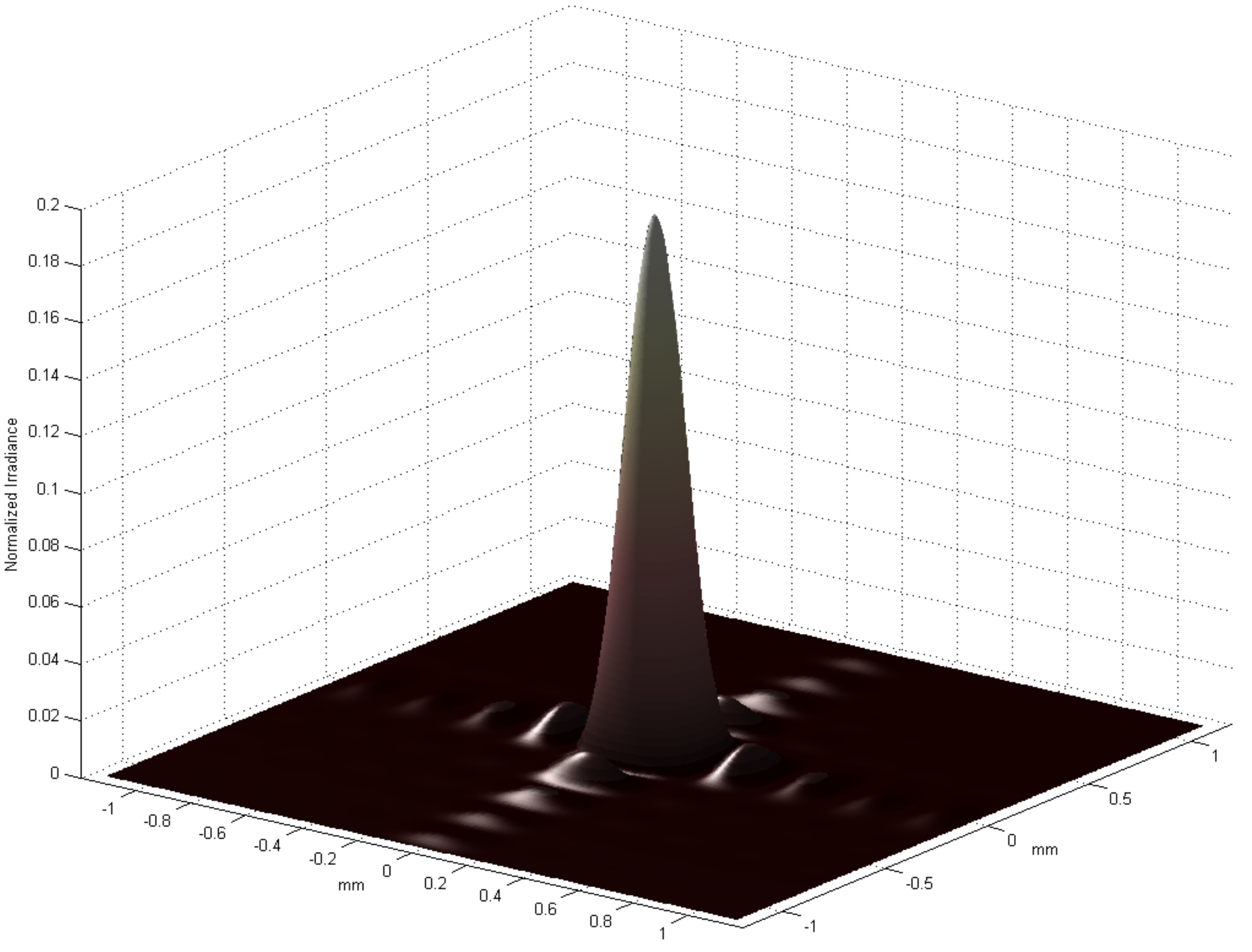}
   \label{fig:diffraction4}
 }
 \subfigure[PSF matrix]{
   \includegraphics[width=3cm] {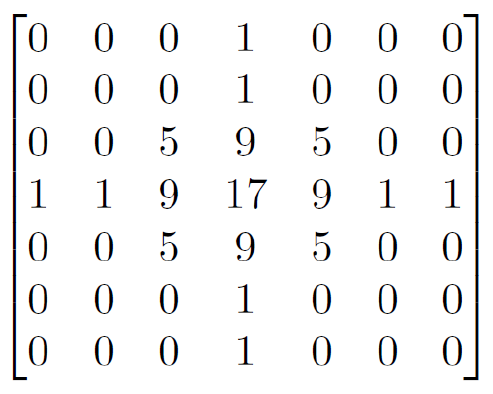}
   \label{fig:diffraction5}
 }

\caption{\textbf{(a)-(c)} Illustration how diffraction by the sensing pattern is modelled as a linear acquisition process through convolution on the CS matrix. \textbf{(a)} Without diffraction the light transmitted by $\mathcal{M}_{i,j}$ is all projected on the correct detector pixel. \textbf{(b)} The transmitted light is diffracted and spread as a PSF over multiple detector pixels. \textbf{(c)} The diffraction is modelled as a PSF on the modulator instead of on the detector, this results in a linear projection model as in the ideal case but with the measurements as acquired in (b). \textbf{(d)} PSF calculated over an area corresponding to 23$\times$23 pixels based on the dimensions of the presented architecture \textbf{(e)} Distribution in percent of total irradiance in the central part of the PSF (rounded for display purposes only). }
\label{fig:diffraction}
\end{figure*}

The ray representation of light used in the previous sections is only accurate for large scales. At small scales, close to the wavelength of the radiation, diffraction becomes a prominent phenomenon \cite{optics:Goodman}. In our architecture, diffraction will mostly be noticeable in the modulator pixels, since all other optics involved require an aperture large enough to cover \emph{all} modulator pixels; they can safely be neglected when comparing to the aperture of a modulator pixel. The \emph{point spread function} (PSF) describing the diffraction {\color{black}of a modulator pixel} is estimated using Fourier optics to calculate the expected image of a point with an incoherent imaging system, as described in {\color{black}section 7.3.3} of \cite{optics:Voelz}. {\color{black}In a real system, however, the effects of diffraction can be measured more accurately by acquiring the response of single image points on the entire modulator pattern.} Fig. \ref{fig:diffraction1} illustrates {\color{black}the behaviour} without diffraction and \ref{fig:diffraction2} with diffraction.

{\color{black}Using this model} the diffraction is a convolution ($*$) of our measurements from the linear projection model (\ref{eqn:DMI}) by the PSF $h$. Since convolution has commutative and associative properties we can write $D_{diff}=D \ast h=(MI) \ast h=(M \ast h) I$ 
(Note that this notation is simplified as the convolution need to be adapted on the non-vectorized forms).
The right hand side provides a method to solve the deconvolution problem directly in the CS reconstruction stage by using the sensing matrix $A=M \ast h$ and $y=D_{diff}$ in (\ref{eqn:CSsensing}) to recover the original image. This is illustrated in Fig. \ref{fig:diffraction3}. {\color{black}Instead of showing that the RIP condition still holds, we have simulated the acquisition and reconstruction of multiple test images.}

\vspace{-0.3cm}
\begin{table}[ht]
\caption{Comparison of normalized reconstruction error as $MSE/MSE_{128\times128}$ mean (standard deviation)} \vspace{-0.2cm} 
\centering 
\begin{tabular}{l p{1.1cm} c c c c c} 
\hline \hline  
Image & Classic 64$\times$64 & CI 4069 & A 64$\times$64 & B 64$\times$64\\ [0.2ex] 
\hline \hline
R            	& 2.014 (0)& 1.692 (0.003)& 1.581 (0.007)& 1.576 (0.083)\\
Lena		& 2.120  (0)& 2.088 (0.030)& 1.792 (0.038)& 1.750 (0.046)\\
Birds		& 2.064  (0)& 1.849 (0.017)& 1.809 (0.026)& 1.833 (0.060)\\
Monarch  	& 2.160  (0)& 2.201 (0.039)& 1.785 (0.034)& 1.763 (0.036)\\
Boat 		& 1.886  (0)& 2.079 (0.026)& 1.782 (0.023)& 1.760 (0.020)\\
Peppers 	& 2.139  (0)& 1.930 (0.020)& 1.696 (0.034)& 1.682 (0.034)\\
Goldhill 	& 1.833 (0)& 2.282 (0.030)& 1.847 (0.032)& 1.765 (0.023)\\
Couple 	& 1.730 (0)& 2.151 (0.020)& 1.848 (0.028)& 1.779 (0.017)\\ [0.1ex] 



\hline 
\end{tabular}
\label{table:MSE} 
\end{table}
\begin{figure*}[!ht]
\includegraphics[width=5.75in]{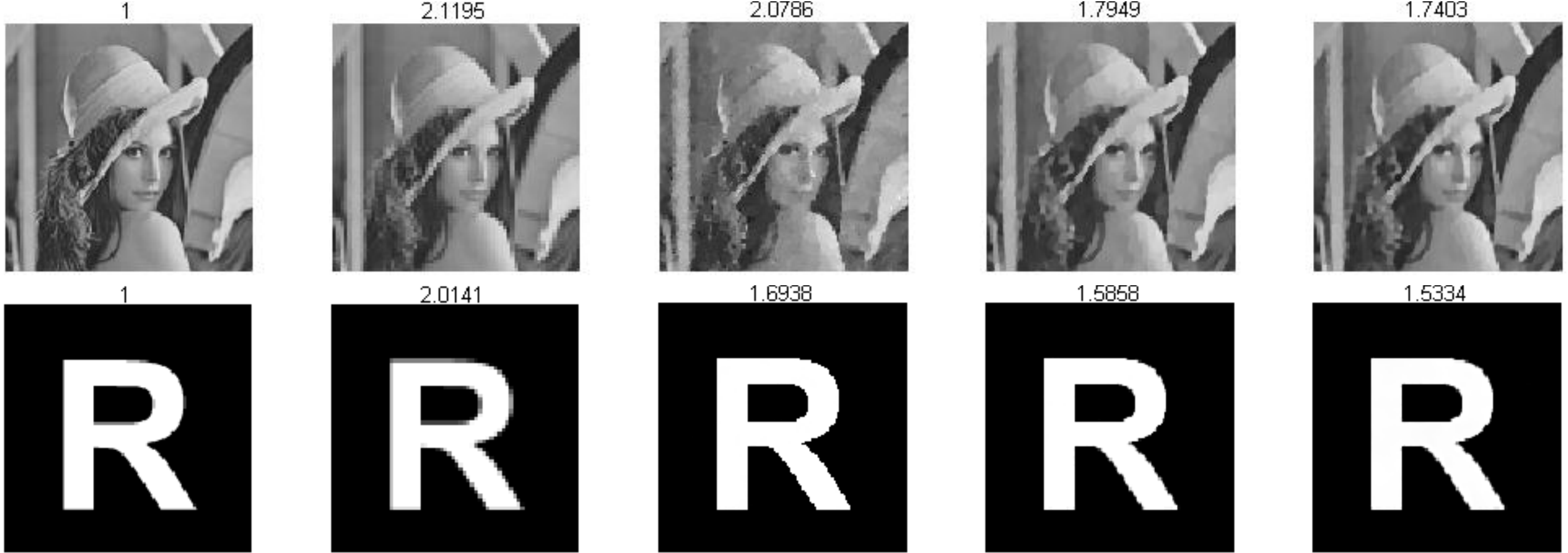}
\centering
\setlength{\abovecaptionskip}{0pt} 
\caption{Reconstructed images. Column 1: Classic camera 128$\times$128. 2: Classic camera 64$\times$64. 3: Sequential CI camera 4096 measurements. 4: Parallel CI 4096 measurements, double shifts. 5: Parallel CI 4096 measurements, double detector pixel size. $MSE/MSE_{128\times128}$ is indicated above each image.}
\label{fig:samples}
\end{figure*}
\vspace{-0.4cm}
\section{Experiments of Acquisition and Reconstruction}
Our simulations are based on a system with realistic dimension limitations on the size of available lenses, SLMs and sensor arrays. Because of the ability to recover (to an unknown degree) the unconvolved image despite a large PSF, different dimensions were tested rather than basing our design on minimizing the PSF. After an extensive comparison of different alternatives in terms of SLM resolution and size as well as projection distance, an SLM of size 25.6 mm $\times$ 25.6 mm with 0.1 mm pixels at a projection distance of 60 mm proved to give the best results. The following simulations are based on these dimensions. We also limit the simulations to recovery with 1/4 of the measurements classically required by the Nyquist-rate. For an image resolution of 128$\times$128 pixels, this requires a detector resolution of 64$\times$64 pixels. Both architectures uses a $0/1$ sensing matrix and converts the measurements as acquiring $I_{total}$ separately (A) or by deriving $I_{total}$ from the measurements (B). The PSF was calculated on the above given dimensions and considering incoherent light at a wavelength of 400 nm; the resulting PSF is shown in Fig. \ref{fig:diffraction4}. As a comparison we simulate a normal digital camera with a resolution of 128$\times$128 pixels and one with 64$\times$64 pixels and a CI camera using independent measurements, such as a single pixel camera. All these cameras are considered to have a negligible PSF. Image acquisition with the classic cameras is simulated by averaging all original image pixels within the regions of the cameras larger pixels (4$\times$4 and 8$\times$8 respectively{\color{black}, all original images are 512$\times$512 pixels}). The same procedure is performed on the sensing matrix for the CI cameras  {\color{black} and the PSF was applied on the measurements of (A) and (B)}. To recover the images of the CI reference camera and our parallel variants, TVAL3 v1.0 \cite{CS:Li} was used. {\color{black} The simulations are programmed in MATLAB and} all cameras are simulated without noise.

We have compared the image quality of 8 test images based on the mean square error (MSE) with respect to the original images. Fig. \ref{fig:samples} shows the reconstruction of two of the test images after simulated acquisition with all architectures and Tab. \ref{table:MSE} shows the reconstruction errors of 8 test images, normalized by the errors of the reference camera with 128$\times$128 pixels. Presented values are averages of 25 reconstructions using different random generations of $M$, with the standard deviation in parenthesis. The first column shows the error of a digital camera using the same amount of measurements (pixels) as the CI cameras, the CI cameras all show comparable results and architectures A and B both outperforms the digital camera on most images. The sequential CI camera sometimes show slightly worse results but still shows a significant improvement on ``R".

\vspace{-0.2cm}
\section{Conclusion}
\vspace{-0.1cm}
In this paper we show that image reconstruction is possible through parallel acquisition of measurements subjected to diffraction with comparable results to a CI camera with independent, sequential measurements which require a significantly longer acquisition time. We are currently assembling a hardware prototype of this architecture, and we will report experimental results in a future paper.
\vspace{-0.2cm}
\section{Acknowledgment}
\vspace{-0.1cm}
This work is supported by the European Research Council under the European Community’s Seventh Framework Programme (FP7/2007-2013) / ERC Grant agreement n.279848.
\end{document}